\PassOptionsToPackage{dvipsnames}{xcolor}
\documentclass{article} 
\usepackage{iclr2025_conference,times} 

\usepackage{url}

\usepackage{graphicx}
\usepackage{booktabs}

\usepackage{amsmath}
\usepackage{amssymb}
\usepackage{tabularx}
\usepackage{bbold}
\usepackage{caption}
\usepackage{comment}
\usepackage{pifont}
\usepackage{wasysym}
\usepackage{graphicx}

\usepackage{color}
\usepackage{colortbl}
\usepackage{multirow}

\usepackage{float}
\usepackage{wrapfig}
\usepackage{lipsum} 
\usepackage{mdframed}
\usepackage{tcolorbox}
\usepackage{wrapfig,lipsum}
\usepackage{hyperref}

\newcommand{\eg}[1]{\textit{e.g.,}}
\newcommand{\ie}[1]{\textit{i.e.,}}
\newcommand{\etal}[1]{\textit{et al.}}

\definecolor{weakorange}{RGB}{255,230,200} 
\definecolor{weakgray}{RGB}{240,240,240} 
\definecolor{colbest}{rgb}{0.1, 0.6, 0.1}
\definecolor{colworst}{rgb}{0.75, 0, 0}
\definecolor{oursrow}{HTML}{EBF5F8}
\definecolor{rebuttal}{rgb}{1.0,0.0,0.0}

\usepackage{xcolor}   

\usepackage{hyperref}  
\hypersetup{
  colorlinks  = true,
  citecolor   = Blue,
  pdfborder   = {0 0 0}
}

\renewcommand{\href}[2]{%
  \orghref{#1}{\textcolor{NavyBlue}{\uline{#2}}}%
}
\makeatother

\title{\textsc{Saner}: Annotation-free Societal Attribute Neutralizer for Debiasing CLIP}

\author{%
  Yusuke Hirota\textsuperscript{1,2}\thanks{Work done as an intern at NVIDIA.}, 
  Min-Hung Chen\textsuperscript{1}, 
  Chien-Yi Wang\textsuperscript{1}, 
  Yuta Nakashima\textsuperscript{2}, \\
  \textbf{Yu-Chiang Frank Wang}\textsuperscript{1,3}, 
  \textbf{Ryo Hachiuma}\textsuperscript{1} \\
  \textsuperscript{1}NVIDIA \quad
  \textsuperscript{2}Osaka University \quad
  \textsuperscript{3}National Taiwan University \\
  \texttt{\{y-hirota,nakashima\}@is.ids.osaka-u.ac.jp} \\
  \texttt{\{minhungc,chienyiw,frankwang,rhachiuma\}@nvidia.com}
}

\iclrfinalcopy 
\begin{document}

\maketitle

\begin{abstract}
Large-scale vision-language models, such as CLIP, are known to contain societal bias regarding protected attributes (\eg, gender, age). This paper aims to address the problems of societal bias in CLIP. Although previous studies have proposed to debias societal bias through adversarial learning or test-time projecting, our comprehensive study of these works identifies two critical limitations: 1) \textit{loss of attribute information} when it is explicitly disclosed in the input and 2) \textit{use of the attribute annotations} during debiasing process. To mitigate societal bias in CLIP and overcome these limitations simultaneously, we introduce a simple-yet-effective debiasing method called \textbf{SANER} (\underline{s}ocietal \underline{a}ttribute \underline{n}eutraliz\underline{er}) that eliminates attribute information from CLIP text features only of \textit{attribute-neutral} descriptions. Experimental results show that SANER, which does not require attribute annotations and preserves original information for \textit{attribute-specific} descriptions, demonstrates superior debiasing ability than the existing methods. 
\footnote{Project page: \url{https://rebnej.github.io/saner-clip.github.io/}}
\end{abstract}

\section{Introduction}
\label{sec:intro}

Large-scale vision-language models (VLMs), such as CLIP \citep{radford2021learning}, have demonstrated a remarkable capability in multi-modal understanding \citep{Luddecke2022CVPR,tewel2022zerocap} and generation \citep{rombach2022high,tao2023galip,yamazaki2023vltint}, being trained with million-scale image-text pairs. Utilizing these VLMs, recent vision models have achieved significant performance enhancements across a wide range of computer vision tasks (\eg, captioning \citep{mokady2021clipcap,yamazaki2022vlcap,li2023decap} and object detection \citep{Li2022CVPR,Zhong2022CVPR}), without the necessity for task-specific training \citep{shen2021much}.

Despite the success, several works have identified societal bias regarding demographic attributes, such as gender and age, in these VLMs \citep{wolfe2022markedness,hausladen2024causal,alabdulmohsin2023clip}, potentially causing unfair or prejudicial decisions by models. \citet{hall2023vision} conducted audits on performance disparity, particularly with respect to gender, and revealed gender-dependency of the CLIP performance. \citet{qiu2023gender} also demonstrated that adopting CLIP for caption evaluation tends to favor gender-stereotypical sentences (\eg, preferring ``A woman is cooking'' over ``A man is cooking'' for images depicting men), highlighting the inherent gender bias. These findings underscore the importance of addressing bias in VLMs.

Some studies have proposed mitigating societal bias in VLMs \citep{berg2022prompt,seth2023dear,dehdashtian2024fairvlm,chuang2023debiasing}. \textit{Adversarial debiasing} \citep{berg2022prompt,seth2023dear,dehdashtian2024fairvlm} fine-tunes CLIP to lessen leakage of protected attributes\footnote{We refer to any demographic variables, like age and gender, as \textit{protected attribute} (or \textit{attribute} in short), based on which a model's decisions should not be made.} into the features, while \textit{projection-based debiasing} \citep{chuang2023debiasing} removes the protected attribute encoded in CLIP features at the inference phase. Our holistic review of these pioneering works (Sec.~\ref{sec:analysis}), though, identifies the following potential drawbacks or controversies in their design choices.

\begin{figure}[t]
  \centering
  \includegraphics[clip, width=0.95\textwidth]{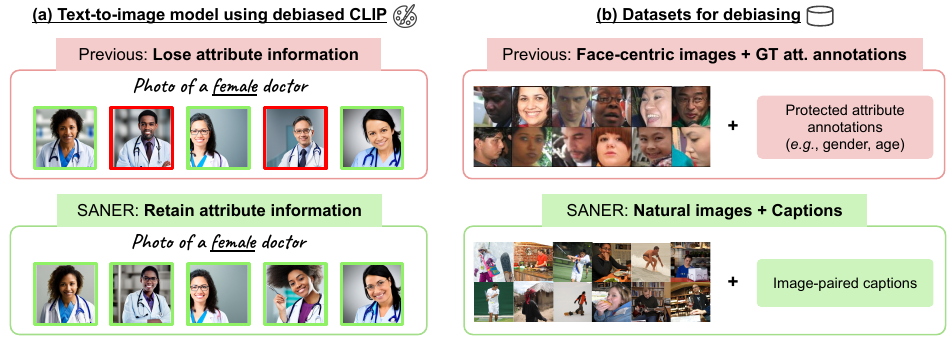}
  \vspace{-5pt}
  \caption{Our debiasing method, SANER, overcomes the limitations in existing methods: (a) attribute information is retained after debiasing, and (b) protected attribute annotations are not required for debiasing.  }
  \label{fig:first}
  \vspace{-12pt}
\end{figure}

\noindent \textbf{Loss of attribute information explicitly disclosed in the input.} Some methods aim to completely remove attribute information by decorrelating the attribute and the features \citep{dehdashtian2024fairvlm} or by squashing the subspace associated with the attribute \citep{chuang2023debiasing}, even when the attribute is explicitly disclosed. This choice can limit the generalizability of a VLM's features to a spectrum of downstream tasks (\eg, Stable Diffusion~\citep{rombach2022high} generates male images for text prompt ``a female doctor'' when encoded with debiased CLIP as shown in Fig. \ref{fig:first} (a) and Sec. \ref{sec:exp-att-info}), while it works for attribute-agnostic downstream tasks \citep{krause20133d}. 

\noindent \textbf{Use of the attribute annotations.} Adversarial debiasing methods  \citep{berg2022prompt,seth2023dear,dehdashtian2024fairvlm} require protected attribute annotations, as provided in FairFace~\citep{karkkainen2021fairface}, for fine-tuning (Fig. \ref{fig:first} (b)). Datasets with attribute annotations are still scarce, partly because the annotation process needs ethical considerations \citep{andrews2023ethical}, limiting their applicability. This dataset scarcity also causes the limited diversity of images and text descriptions used for fine-tuning the VLM, potentially inducing \textit{overfitting}.

This paper presents a simple-yet-effective debiasing approach for CLIP, called \textbf{SANER} (\underline{s}ocietal \underline{a}ttribute \underline{n}eutraliz\underline{er}), that simultaneously overcomes the aforementioned limitations. 
Specifically, SANER trains a debiasing layer (\ie, a multilayer perception) to amend CLIP text feature vectors of \textit{attribute-neutral} descriptions, given by \textbf{attribute neutralization}, such that they are equidistant to those of \textit{attribute-specific} descriptions using \textbf{annotation-free debiasing loss}. 
With this, only feature vectors for attribute-neutral descriptions are debiased, whereas the attribute-specific ones retain the original information. Attribute-specific descriptions for all possible attribute groups\footnote{\textit{Attribute group} is a class in a protected attribute (\eg, \textit{female} and \textit{male} in gender).} can be easily augmented by modifying the attribute-specific words in the original descriptions, directing the training without attribute annotations.

\textbf{Contribution}. Thanks to our annotation-free debiasing pipeline, SANER is designed to be compatible with any dataset of image-text pairs, such as COCO \citep{lin2014microsoft}. This provides denser guidance for training the debiasing layer compared to the existing methods. Moreover, SANER does not require retraining the CLIP model itself, accessing its original training data, or retraining downstream tasks (\eg, text-to-image generation) when applying the debiased CLIP. Experiments on both discriminative and generative tasks (\ie, text-to-image retrieval \citep{geyik2019fairness} and text-to-image generation \citep{rombach2022high}) show that SANER can mitigate gender, age, and racial biases of CLIP. Moreover, we demonstrate that SANER outperforms the existing methods \citep{berg2022prompt,chuang2023debiasing}, showing that SANER leads to less attribute-dependency of the downstream performance while overcoming the limitations in existing methods. 

\section{Review: Existing Debiasing Methods}
\label{sec:analysis}

Several debiasing approaches for CLIP have been introduced, broadly categorized into two main types: \textit{adversarial debiasing} \citep{berg2022prompt,seth2023dear,dehdashtian2024fairvlm} and \textit{projection-based debiasing} \citep{chuang2023debiasing}. This section conducts an in-depth analysis of these existing debiasing strategies, highlighting their respective limitations.

\vspace{5pt}
\noindent
\textbf{Notation. }
Let $\mathcal{D}$ denote a dataset, each of whose sample is quadruple $(v, t, a, d)$, 
where $v$ is an image, 
$t$ is a text description, 
$a \in \mathcal{A}$ is a protected attribute annotation from set $\mathcal{A}$ of all attribute groups, and $d$ is the ground-truth annotation for a downstream task (if any). The CLIP text and image encoders, denoted by $f_\text{t}(t) \in \mathbb{R}^K$ and $f_\text{v}(v) \in \mathbb{R}^K$, respectively, take $t$ and $v$ as input and generate corresponding feature vectors in a common space. 

\subsection{Adversarial debiasing}
Adversarial debiasing \citep{berg2022prompt,seth2023dear,dehdashtian2024fairvlm} aims to eliminate protected attribute information in the CLIP features. Specifically, an adversarial classifier is employed to predict and remove protected attribute $a$ from CLIP features. 

\vspace{5pt}
\noindent
\textbf{Prompt tuning-based debiasing} \citep{berg2022prompt} 
proposes to use learnable tokens to reduce attribute leakage through the similarity between an image and a set of pre-defined textual concepts. Concretely, for a set $\mathcal{C}$ of pre-defined concepts (\ie, phrases) that are supposed to be attribute non-specific (\eg, \textit{smart} and \textit{attractive}), a sequence $l$ of $k$ learnable tokens are prepended to the sentence template $t_c$ with concept $c \in \mathcal{C}$ (\eg, $t_c = \text{``A photo of a smart person''}$ for $c = \textit{smart}$) to obtain $t_c' = [l, t_c]$, where $[\cdot, \cdot]$ represents sequence concatenation. Then, $t'_c$ and arbitrary $v \in \mathcal{D}$ is fed into the CLIP encoders to compute similarity $s_c(v)$ by 
\begin{equation}
    s_c(v) = f_\text{v}(v)^\top f_\text{t}(t_c').  
\end{equation}
Let $s(v) \in \mathbb{R}^{|\mathcal{C}|}$ 
denote a vector, each of whose elements is the similarity score $s_c(v)$ for a concept in $\mathcal{C}$. Due to the attribute non-specificity of concepts in $\mathcal{C}$, $s(v)$ should not correlate with attribute $a$ of $v$. However, the CLIP text encoder can embed $a$ into $s(v)$ due to bias, allowing an attribute classifier to predict $a$. We denote the probability of being $a$ given $s(v)$ (or a prediction score of the attribute classifier) by $m_a(s(v))$.   
Prompt tuning-based debiasing \citep{berg2022prompt} uses $m_a$ for adversarial loss, given by 
\begin{equation}
    \mathcal{L}_\text{adv} = - \sum_{v \in \mathcal{D}}\log m_a(s(v)).
\end{equation}
Minimizing $\mathcal{L}_\text{adv}$ with respect to $l$ reduces attribute leakage through $s(v)$. 
A contrastive loss between image and text features is also used for regularization. 

The experiments \citep{berg2022prompt} showed that this method could effectively reduce attribute leakage through $f_\text{t}(t)$, but the limited number of concepts\footnote{Their experiments used $10$ concepts.} may limit downstream tasks that enjoy the debiased features because as $l$ is learned only through a sparse set of concepts. Additionally, attribute annotations are necessary for the adversarial loss, resulting in the exclusive use of face-centric image datasets (\eg, FairFace \citep{karkkainen2021fairface}) as $\mathcal{D}$.

\vspace{5pt}
\noindent
\textbf{Additive Residual Learner (ARL)} \citep{seth2023dear}
is designed to remove attribute information from CLIP image features. This method assumes that a debiasing layer\footnote{A fully-connected layer is used.} $r$ can identify a vector to additively amend attribute-neutral image feature vector $\delta(v)$, \ie, 
\begin{equation}
    \delta(v) = f_\text{v}(v) - r(f_\text{v}(v)).
\end{equation}
Similarly to \citep{berg2022prompt}, an adversarial classifier is trained to predict $a$ from $\delta$ with adversarial loss
\begin{equation}
    \mathcal{L}_\text{adv} =  -\sum_{v \in \mathcal{D}} \log  m_a(\delta(v)).
\end{equation}
The reconstruction loss between $f_\text{v}(v)$ and $\delta(v)$ regularizes training to preserve the original features.

This method shares a common limitation with prompt tuning-based debiasing \citep{berg2022prompt}, notably requiring attribute annotations. Another limitation is that it tries to remove attribute features even when attributes of people in images are explicitly disclosed (\ie, when the person is depicted in an image). Consequently, debiased CLIP is ignorant of protected attributes.

\vspace{5pt}
\noindent
\textbf{Mapper} \citep{dehdashtian2024fairvlm} 
aims to reduce spurious correlations between attributes $a$ and task label $d$ in $\mathcal{D}$. It applies mappings $f_\text{v}'$ and $f_\text{t}'$ to image and text features, respectively, as $x_\text{v}(v) = f_\text{v}'(f_\text{v}(v))$ and $x_\text{t}(t) = f_\text{t}'(f_\text{t}(t))$ for mitigating dependence on $a$. The adversarial loss is computed using a dependence measure $\text{Dep}(\cdot, \cdot)$ to quantify statistical dependence between features as:
\begin{equation}
    \mathcal{L}_\text{adv} = -\text{Dep}(x_\text{v}(v), a) - \text{Dep}(x_\text{t}(t), a).
\end{equation}
These mapping functions are also trained to maximize the statistical dependence between the features after the mapping and task label $d$, \ie,
\begin{equation}
\text{Dep}(x_\text{v}(v), d) + \text{Dep}(x_\text{t}(t), d),
\end{equation}
to retain the predictive power on the downstream task while reducing bias.

Similar to \citep{berg2022prompt,seth2023dear}, Mapper relies on attribute annotations. Moreover, it is designed only to address the spurious correlations between the attribute and task labels for a specific task but not for different tasks.

\subsection{Projection-based debiasing}

Projection-based debiasing \citep{chuang2023debiasing} projects CLIP text feature vectors into the orthogonal complement of the space spanned by a set of CLIP text feature vectors that pertain to the protected attribute. Specifically, let $\mathcal{U}$ denote a set of text descriptions with the target attribute (\eg, $\text{``A photo of a $w$''} \in \mathcal{U}$, where $w \in \{\text{``woman'', ``man''}\}$ for binary gender), and $U$ be a matrix each of whose column vectors is $f_\text{t}(u)$ with $u \in \mathcal{U}$. The projection matrix $P$ into the orthogonal complement for $U$ is given by
\begin{equation}
    P = I - U(U^\top U)^{-1}U^\top
\end{equation}
where $I$ is the identity matrix. $P$ can project a CLIP text feature vector $f_\text{t}(t)$ for a text description $t$ into the orthogonal complement by $Pf_\text{t}(t)$. This process removes attribute information by projecting features into the space orthogonal to attribute-specific directions.

Unlike adversarial debiasing, which requires training by gradient descent update, $P$ has a closed-form solution and is computed in the inference phase. However, as with ARL, this method also eliminates attribute information even from descriptions with explicit attributes (\eg, ``A photo of a female doctor'').

\subsection{Summary of the challenges} 

The existing debiasing methods, including adversarial and projection-based, reveal several challenges: 1) \textbf{Loss of attribute information} (ARL and projection-based) even with explicit attribute description narrows down the utility of the debiased CLIP. For example, a text-to-image generative model with gender-debiased CLIP features may not properly depict explicitly specified gender (as shown in Sec. \ref{sec:exp-att-info}), 
2) \textbf{Dependency on attribute annotations} (prompt tuning, ARL, and Mapper) constrains the range of datasets that can be utilized, often necessitating the use of face-centric image datasets (\eg, FairFace), as opposed to more diverse, natural image datasets (\eg, COCO). 

\section{Societal Attribute Neutralizer (SANER)}
\label{sec:method}

Our method for debiasing CLIP features, SANER, addresses the limitations of the existing methods identified in Section \ref{sec:analysis}. Notably, SANER 1) retains attribute information in cases where the person's attributes are explicitly described and 2) eliminates the reliance on attribute annotations, allowing the use of any image-text dataset for training the debiasing layer.

SANER comprises 1) \textbf{attribute neutralization}, which eliminates protected attribute information from input text (Section \ref{sec:neut}); 2) \textbf{feature modification}, which removes attribute information from the CLIP text features by amending them with a debiasing layer (Section \ref{sec:modification}); 3) \textbf{attribute annotation-free debiasing loss}, ensuring the features are not biased towards any attribute group $g \in \mathcal{A}$ (Section \ref{sec:debiasing}); and 4) \textbf{regularization losses}, which preserve the original CLIP features and the alignment between image and text features (Section \ref{sec:regul}).  

\begin{figure}[t]
  \centering
  \includegraphics[clip, width=0.92\textwidth]{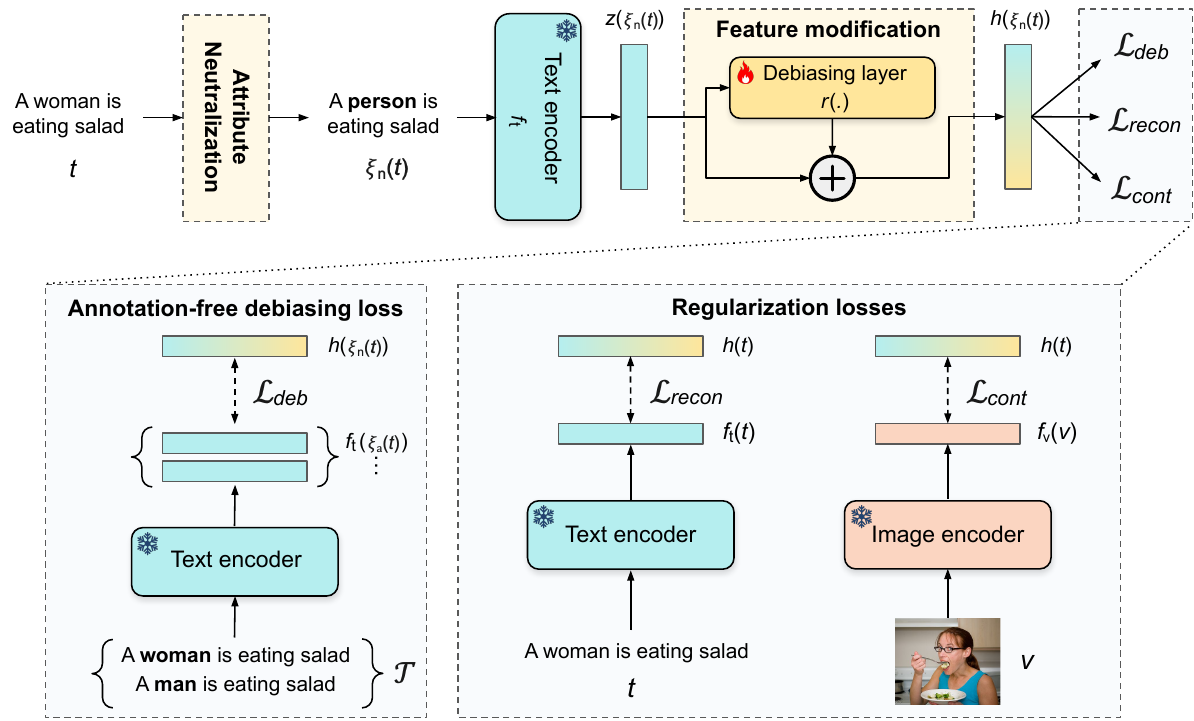}
  \vspace{-5pt}
  \caption{An overview of SANER, exemplified by binary gender. SANER neutralizes attribute-specific text (\eg, ``woman'' $\rightarrow$ ``person''), modifies features via debiasing layer, and uses three losses for debiasing: $\mathcal{L}_\text{deb}$ for attribute neutralization, $\mathcal{L}_\text{recon}$ for feature preservation, and $\mathcal{L}_\text{cont}$ for image-text alignment.}
  \label{fig:method}
  \vspace{-12pt}
\end{figure}

Figure \ref{fig:method} shows an overview of SANER. We train the debiasing layer for feature modification over an arbitrary dataset $\mathcal{D} = \{(v, t)\}$ of image $v$ and text description $t$ (\eg, image caption, alt text) pairs, which does not provide attribute annotation $a$ as well as target task label $d$.

\subsection{Attribute neutralization}
\label{sec:neut}

We first modify text description $t \in \mathcal{D}$ that contain person-related words,\footnote{Person-related words encompass terms that reference individuals (\eg, \textit{person}, \textit{girl}, \textit{man}). The complete list is in the appendix.} to remove attribute-specific words. Taking binary gender as a protected attribute,\footnote{Following prior research \citep{berg2022prompt,seth2023dear,dehdashtian2024fairvlm,chuang2023debiasing,zhao2017mals,garcia2023uncurated,burns2018women}, we focus on the binary gender but recognize the importance of inclusivity. SANER applies to non-binary genders.} \ie, $\mathcal{A} = \{\texttt{female}, \texttt{male}\}$, as example, the text description
\begin{quote}
\centering
$t = \text{``A woman is eating salad.''}$
\end{quote}
contains attribute information (\ie, \textit{woman}). We replace attribute-specific terms\footnote{We use gender words defined in \citep{hirota2023model}.} with the attribute-neutral ones to obtain an attribute-neutral text:
\begin{quote}
\centering
$\xi_\text{n}(t) = \text{``A \underline{person} is eating salad.''}$
\end{quote}
where $\xi_\text{n}$ denotes a function for attribute neutralization. 
Neutralization can be done for other attributes, such as age.\footnote{Examples for the race attribute are in the appendix.} We remove age-specific terms\footnote{We define age-specific terms. The list is in the appendix.} (\eg, \textit{young} and \textit{senior}) in text descriptions, for instance, ``A young woman is eating salad'' $\rightarrow$ ``A woman is eating salad''. 
In contrast to the previous approach~\citep{chuang2023debiasing}, which is optimized not to predict the attribute information from the original description $t$, we target the attribute-neutral descriptions $\xi_\text{n}(t)$ to preserve the attribute information in the features of attribute-specific descriptions.

\subsection{Feature modification}
\label{sec:modification}

CLIP text features $z(\xi_\text{n}(t)) = f_\text{t}(\xi_\text{n}(t))$ after attribute neutralization can still convey the protected attribute information due to CLIP's bias. To remove such bias, we append a learnable debiasing layer $r$ on top of $f_\text{t}$, inspired by recent CLIP fine-tuning techniques \citep{seth2023dear,gao2024clip}. Neutralized $t$'s debiased feature $h(\xi_\text{n}(t))$ is given by
\begin{equation}
    h(\xi_\text{n}(t)) = z(\xi_\text{n}(t)) + r(z(\xi_\text{n}(t))).
\end{equation}

\subsection{Attribute annotation-free debiasing loss}
\label{sec:debiasing}

To train $r$ to extract attribute information from CLIP features without attribute annotations, we create a set $\mathcal{T}$ of attribute-specific descriptions for $t \in \mathcal{D}$ and for $g \in \mathcal{A}$, \ie, $\mathcal{T} = \{\xi_g(t)|t \in \mathcal{D}, g \in \mathcal{A}\}$, where $\xi_g(t)$ generates a description for each attribute group $g$ from $t$. For binary gender, this involves generating descriptions with female- and male-specific words. For instance, from the text description, ``A woman is eating salad.'', we generate two sentences with female and male attributes:
\begin{quote}
\centering
    A \underline{woman} is eating salad. \\
    A \underline{man} is eating salad.
\end{quote}

The debiasing loss trains $r$ such that $h(\xi_\text{n}(t))$ is equidistant from $f_\text{t}(\xi_g(t))$ for all attribute groups in $\mathcal{A}$, ensuring an impartial representation across the spectrum of attribute groups. We implement this loss as the standard deviation of the cosine similarity between $h(\xi_\text{n}(t))$ and $f_\text{t}(\xi_g(t))$. Let $s_g(t)$ denote the similarity, \ie,
\begin{align}
    s_g(t) = \frac{h(\xi_\text{n}(t))^\top f_\text{t}(\xi_g(t))}{\|h(\xi_\text{n}(t))\| \; \|f_\text{t}(\xi_g(t))\|}.
\end{align}

The debiasing loss $\mathcal{L}_\text{deb}$ is defined as
\begin{equation}
\label{eq:deb_loss}
    \mathcal{L}_\text{deb} = \sqrt{\frac{1}{|\mathcal{D}|}\sum_{t \in \mathcal{D}} (s_g(t)-\bar{s}(t))^2},
\end{equation}
where $\bar{s}(t) = \sum_{g \in \mathcal{A}} s_g(t) / |\mathcal{A}|$. A lower standard deviation means $s_g$ is close to $\bar{s}$, leading to $h(\xi_\text{n}(t))$ being equidistant to $f_\text{t}(\xi_g(t))$ for all $g \in \mathcal{A}$.  Notably, this debiasing loss can be computed without attribute annotations.

\subsection{Regularization losses}
\label{sec:regul}

Applying the debiasing loss alone significantly changes original CLIP features, thereby losing semantics \citep{berg2022prompt,seth2023dear}. To maintain the alignment of resulting image-text features, we utilize reconstruction loss \citep{seth2023dear} and contrastive loss \citep{radford2021learning}. Reconstruction loss $\mathcal{L}_\text{recon}$ is the mean squared error between $f_\text{t}(t)$ and $h(t)$. Contrastive loss $\mathcal{L}_\text{cont}$ aims to minimize the negative log-likelihood of input image-caption pairs, $f_\text{v}(v)$ and $f_\text{t}(t)$, in comparison to negative ones. Note that the original description $t$ is used for regularization losses.

\subsection{Training and inference} 
The overall loss $\mathcal{L}$ is given by:
\begin{equation}
\label{eq:our_loss}
    \mathcal{L} = \alpha \mathcal{L}_\text{deb} + \beta \mathcal{L}_\text{recon} + \gamma \mathcal{L}_\text{cont},
\end{equation}
where $\alpha$, $\beta$, and $\gamma$ are the hyperparameters to weight respective losses. 

During inference, we apply the trained debiasing layer $r$ and use the modified text features $r(f_\text{t}(t))$ as the CLIP text features.

\section{Experiments: Text-to-Image Retrieval}
\label{sec:exp1}
Following previous studies \citep{berg2022prompt,seth2023dear,dehdashtian2024fairvlm,chuang2023debiasing}, we evaluate SANER on the text-to-image retrieval task regarding gender, age, and racial biases. 
Further analysis, such as the ablation study of the loss components, is in the appendix. 

\subsection{Experimental settings}

\noindent \textbf{Evaluation metric. }
We employ the \textbf{MaxSkew} metric \citep{geyik2019fairness}, utilized in the previous studies~\citep{berg2022prompt,seth2023dear,dehdashtian2024fairvlm,chuang2023debiasing}, to quantify the societal bias in CLIP in the text-to-image retrieval task. MaxSkew measures the disparity between the attribute distribution of $|\mathcal{A}|$ in the top-$k$ retrieved images. Let $\eta_{ak}(q)$ denote the ratio of images labeled with attribute $a$ in the top-$k$ retrieved images. For attribute neutral query $q$, $\eta_{ak}(q)$ should be $1/|\mathcal{A}|$ if the model is unbiased. $\text{MaxSkew@$k$}$ is defined as:
\begin{equation}
    \text{MaxSkew@$k$} = \max_{a \in \mathcal{A}} \log \frac{\eta_{ak}(q)}{1/|\mathcal{A}|}.
\end{equation}
Ideally, $\text{MaxSkew@$k$}$ is 0 but is larger when a model biased.

\noindent \textbf{Evaluation setting.} For the attribute-neutral queries, we use template-based queries such as ``a photo of a $c$ person'', where $c$ is an attribute-neutral concept. Prior work \citep{berg2022prompt} has defined a list of person-related \textbf{adjectives}, such as \textit{clever} and \textit{attractive}, as attribute-neutral concepts. We extend the list to encompass \textbf{occupations} (\eg, \textit{doctor}, \textit{nurse}), and \textbf{activities} (\eg, \textit{cooking} and \textit{cleaning}) for a more comprehensive evaluation. MaxSkew@$k$ is computed per concept.\footnote{The complete lists of the concepts are in the appendix.} 

We also evaluate \textbf{zero-shot image classification accuracy} on ImageNet-1K \citep{imagenet} to ensure that debiasing does not spoil the original CLIP's performance.

\noindent
\textbf{Evaluation datasets. }
We utilize two datasets, FairFace \citep{karkkainen2021fairface} and PATA \citep{seth2023dear}, which consist of images alongside protected attribute annotations (\eg, \textit{female} and \textit{male} for gender) associated with the person in each image. FairFace consists of $10,954$ cropped face-centric images, while PATA contains $4,934$ natural images with a single person. Most debiasing approaches only report the performance on the FairFace dataset, but we additionally employ PATA to evaluate the debias performance on more diverse images.


\noindent
\textbf{Methods for comparison. }
We compare SANER against existing methods, \ie, prompt tuning-based debiasing \citep{berg2022prompt} and projection-based debiasing \citep{chuang2023debiasing}, whose code is publicly available. Unfortunately, we could not reproduce the other methods\citep{seth2023dear,dehdashtian2024fairvlm} since sufficient reproduction details are unavailable.

\noindent
\textbf{Implementation details. }
Following the previous works \citep{berg2022prompt,chuang2023debiasing}, we employ CLIP \citep{radford2021learning} with ViT-B/16 backbone \citep{dosovitskiy2020image} as a target model in our experiments.
We train the debiasing layer (a multilayer perceptron with two linear layers with ReLU activation \citep{nair2010rectified}) appended to the model using $170,624$ image-caption pairs, which is a subset of the COCO training set \citep{lin2014microsoft} with person-related words/phrases (\eg, \textit{person} and \textit{boy}). 
We empirically set $\alpha$, $\beta$, and $\gamma$ to $1.0$, $0.1$, and $0.0001$ (Eq. \ref{eq:our_loss}), respectively, and train for $5$ epochs. Further details are provided in the appendix.

\begin{table}[t]
\renewcommand{\arraystretch}{1.1}
\setlength{\tabcolsep}{5pt}
\scriptsize
\centering
\caption{\textbf{Gender bias}, evaluated by MaxSkew@1000 (scaled by $100$), on FairFace and PATA for the original CLIP (Original), prompt tuning-based debiasing (Prompt), projection-based debiasing (Projection), and our method (SANER). A lower value is better (less gender bias). \textbf{Bold} represents the best across the models.}
\vspace{-9pt}
\begin{tabularx}{0.84\columnwidth}{l r r r r r r r}
\toprule
& \multicolumn{3}{c}{FairFace} &&\multicolumn{3}{c}{PATA}\\ 
\cline{2-4} 
\cline{6-8}
\multirow{-2}{*}{CLIP Model} & \multirow{1.3}{*}{Adjective} & \multirow{1.3}{*}{Occupation} & \multirow{1.3}{*}{Activity} & & \multirow{1.3}{*}{Adjective} & \multirow{1.3}{*}{Occupation} & \multirow{1.3}{*}{Activity}   \\
\midrule
Original \citep{radford2021learning}  & $22.9$ & $33.7$ & $19.5$ && $12.1$ & $18.7$ & $10.7$ \\
\midrule
Prompt \citep{berg2022prompt} & $12.3$ & $29.9$ & $20.0$ && $6.7$ & $16.5$ & $10.2$ \\
Projection \citep{chuang2023debiasing} & $15.4$ & $37.4$ & $15.0$ && $6.4$ & $13.6$ & $5.4$ \\
\cellcolor{oursrow}SANER (Ours)  & \cellcolor{oursrow}$\mathbf{8.9}$ & \cellcolor{oursrow}$\mathbf{14.5}$ & \cellcolor{oursrow}$\mathbf{7.7}$ &\cellcolor{oursrow}& \cellcolor{oursrow}$\mathbf{5.4}$ & \cellcolor{oursrow}$\mathbf{9.5}$ & \cellcolor{oursrow}$\mathbf{3.3}$ \\
\bottomrule
\end{tabularx}
\label{tab:gender-retrieve}
\vspace{-8pt}
\end{table}

\begin{table}[t]
\renewcommand{\arraystretch}{1.1}
\setlength{\tabcolsep}{5pt}
\scriptsize
\centering
\caption{\textbf{Age bias} (left) and \textbf{racial bias} (right), evaluated by MaxSkew@1000 (scaled by $100$), on FairFace. \textbf{Bold} denotes the best across the models. Results on PATA are in the appendix.}
\vspace{-8pt}
\begin{minipage}{0.42\linewidth}
\centering
\begin{tabularx}{\linewidth}{l r r r }
\toprule
 & \multicolumn{3}{c}{FairFace} \\ 
\cline{2-4} 
\multirow{-2}{*}{CLIP Model} & \multirow{1.3}{*}{Adjective} & \multirow{1.3}{*}{Occupation} & \multirow{1.3}{*}{Activity}  \\
\midrule
Original  & $111.1$ & $121.1$ & $113.0$  \\
\midrule
Projection & $107.6$ & $112.8$ & $\mathbf{100.0}$  \\
\cellcolor{oursrow}SANER (Ours) & \cellcolor{oursrow}$\mathbf{96.0}$ & \cellcolor{oursrow}$\mathbf{112.6}$ & \cellcolor{oursrow}$101.9$  \\
\bottomrule
\end{tabularx}
\end{minipage}
\hspace{0.04\linewidth} 
\begin{minipage}{0.42\linewidth}
\centering
\begin{tabularx}{\linewidth}{l r r r }
\toprule
 & \multicolumn{3}{c}{FairFace} \\ 
\cline{2-4} 
\multirow{-2}{*}{CLIP Model} & \multirow{1.3}{*}{Adjective} & \multirow{1.3}{*}{Occupation} & \multirow{1.3}{*}{Activity}  \\
\midrule
Original & $62.2$ & $57.4$ & $68.3$  \\
\midrule
Projection & $56.9$ & $75.3$ & $67.0$  \\
\cellcolor{oursrow}SANER (Ours) & \cellcolor{oursrow}$\mathbf{49.3}$ & \cellcolor{oursrow}$\mathbf{45.7}$ & \cellcolor{oursrow}$\mathbf{46.6}$  \\
\bottomrule
\end{tabularx}
\end{minipage}
\vspace{-11pt}
\label{tab:age-retrieve}
\end{table}

\subsection{Gender bias analysis}
Table \ref{tab:gender-retrieve} presents MaxSkew@1000 on FairFace and PATA for gender bias, showing SANER mitigates bias the most among all methods. In contrast to the existing methods, this tendency is consistent across 1) datasets with different image domains (\ie, face-centric and natural images) and 2) concept types (\ie, adjective, occupation, and activity). For instance, while the prompt tuning-based method \citep{berg2022prompt} fails to mitigate bias for activity concepts on FairFace (\ie, bias is amplified from $19.5$ to $20.0$), SANER significantly mitigates bias from $19.5$ to $7.7$. This verifies the better debiasing performance of SANER compared to the existing methods on diverse concepts and image domains, possibly because SANER is trained with diverse text descriptions (\ie, captions in COCO), which are not constrained like pre-defined concepts required in the previous method \citep{berg2022prompt}.
%

\subsection{Age and racial biases analysis}
Tables \ref{tab:age-retrieve} presents the results of MaxSkew@1000 for age and racial biases. We compare SANER with projection-based debiasing~\citep{chuang2023debiasing}, as the prompt tuning-based method \citep{berg2022prompt} does not provide age and race debiasing variants. Similarly to the results for gender bias, SANER surpasses the existing method across the datasets and the concept types. For example, SANER successfully mitigates the racial bias on FairFace. Meanwhile, the projection-based method amplifies the bias on the occupation concept (\ie, from $57.4$ to $75.3$). These results validate the generalizability of SANER in bias mitigation across the protected attributes.
%

\subsection{Zero-shot image classification}

\begin{wraptable}{R}{0.32\linewidth}
\vspace{-20pt}
\centering
\caption{Accuracy on Image-Net-1K.}
\vspace{-5pt}
\scriptsize
\begin{tabular}{l c r} \toprule
CLIP Model & & Acc. (\%)   \\ \midrule
Original & & $65.4$ \\ \midrule
Prompt & & $64.1$ \\
\cellcolor{oursrow}SANER (Ours) &\cellcolor{oursrow} & \cellcolor{oursrow}$65.2$ \\
\bottomrule
\end{tabular}
\vspace{-12pt}
\label{tab:imagenet}
\end{wraptable}

To verify whether debiasing harms the zero-shot image classification performance of the original CLIP, we evaluate the prompt tuning-based method \citep{berg2022prompt} and SANER on ImageNet-1K in terms of classification accuracy. Projection-based debiasing \citep{chuang2023debiasing} does not apply to this evaluation because zero-shot prompts, such as ``a photo of a car,'' do not necessarily include person-related words.\footnote{Projection-based debiasing requires modified input text to include attribute terms.} 
The results are shown in Tab. \ref{tab:imagenet}, showing that applying SANER maintains classification performance, whereas the performance of the prompt tuning-based method slightly degrades.

\begin{tcolorbox}[colback=yellow!10, colframe=black, boxrule=1pt, arc=1mm]
\textbf{Insight}: Our method, SANER, which does not require attribute annotations, outperforms previous methods—including those that do require annotations—in mitigating gender, age, and racial biases without compromising classification performance.
\end{tcolorbox}

\section{Experiments: Text-to-Image Generation}
\label{sec:exp2}

We also evaluate SANER on text-to-image generation, for which societal bias is actively investigated \citep{bansal2022well,cho2023dall,liu2024scoft}. Specifically, we conduct two experiments from different aspects: 1) gender bias regarding occupations using gender-neutral prompts (Sec. \ref{sec:exp-sd-gender}), and 2) retention of attribute information when prompts explicitly disclose gender (Sec. \ref{sec:exp-att-info}).

\noindent
\textbf{Image generation settings. }
We use Stable Diffusion (SD) v2.1~\citep{rombach2022high} as the text-to-image generation model. The CLIP text encoder in SD is replaced with the debiased one for evaluation. Following \citep{chuang2023debiasing}, we use gender-neutral prompts with specifying occupations to analyze gender bias in generated images. These prompts are derived from the template ``A photo of a $o$'', where $o$ is replaced with specific occupation terms, such as \textit{doctor} and \textit{teacher}.\footnote{The list of the occupations is in the appendix.}
On the other hand, we employ gender-specific prompts to evaluate the capability of attribute information retention. Concretely, a gender term, either \textit{female} or \textit{male}, is added just before the occupation terms, \ie, ``A photo of a \{female/male\} $o$'', to see if generated images specify the gender.
We generate $100$ images for each prompt with SD's default hyperparameters.

\noindent
\textbf{Evaluation metrics. }
For the bias evaluation for the generative task, we use \textbf{statistical parity} (SP) metric that measures the disparity of attribute groups in generated images \citep{teo2024measuring,chuang2023debiasing}. Specifically, we annotate binary gender labels (\ie, $\mathcal{A} = \{\texttt{female}, \texttt{male}\}$) for the generated images with the assistance of human workers.\footnote{Different from previous works that use pre-trained gender classifiers to assign gender labels, we do not use them due to their bias issues \citep{ramaswamy2020debiasing,das2018mitigating,dinan2020multi}.} SP is defined as the difference between the empirical distribution $\kappa_a$ of gender $a$ and uniform distribution, given by:
\begin{equation}
        \text{SP} = \sqrt{\sum_{a \in \mathcal{A}} (\kappa_{a} - 1/|\mathcal{A}|)^2},
\end{equation}
where $\kappa_a = N_a / \sum_{a'} N_{a'}$ with $N_a$ being the number of images annotated as $a$.
For an unbiased text-to-image generation model, SP should be 0 but increases for biased models.

For gender-specific prompts for evaluating gender information retention, we compute the \textbf{accuracy}, \ie, the ratio of the images that contain the same gender as the prompt to all generated images.

\subsection{Gender bias analysis}
\label{sec:exp-sd-gender}

\begin{wraptable}{R}{0.45\textwidth}
\vspace{-12pt}
\centering
\caption{Gender bias (SP) and gender information retention (Accuracy) in images from Stable Diffusion (Original) and the model using projection-based debiased CLIP (Projection) and our debiased CLIP (SANER). 
Results are the mean across occupations. Female and male refer to prompts specifying each gender. \textbf{Bold} indicates the best.
}
\scriptsize
\vspace{-7pt}
\begin{tabular}{l r r r r}
\toprule
& &&\multicolumn{2}{c}{Accuracy $\uparrow$}\\ 
\cline{4-5}
CLIP Model & SP $\downarrow$&& \multirow{1.3}{*}{Female} & \multirow{1.3}{*}{Male}   \\
\midrule
Original  & $0.51$  && $\mathbf{1.00}$  & $\mathbf{1.00}$ \\
\midrule
Projection & $0.47$ && $0.58$ & $0.79$  \\
\cellcolor{oursrow}SANER (Ours) & \cellcolor{oursrow}$\mathbf{0.39}$  &\cellcolor{oursrow}& \cellcolor{oursrow}$\mathbf{1.00}$  & \cellcolor{oursrow}$\mathbf{1.00}$ \\
\bottomrule
\end{tabular}
\vspace{-25pt}
\label{tab:sd-gender}
\end{wraptable} 

Table \ref{tab:sd-gender} summarizes SP scores, which demonstrate that applying SANER to Stable Diffusion notably mitigates gender bias regarding occupations. SANER again outperforms projection-based debiasing one (\ie, $0.39$ for SANER and $0.47$ for projection-based), highlighting the superiority of SANER.

\begin{figure}[t]
  \centering
  \includegraphics[clip, width=0.95\textwidth]{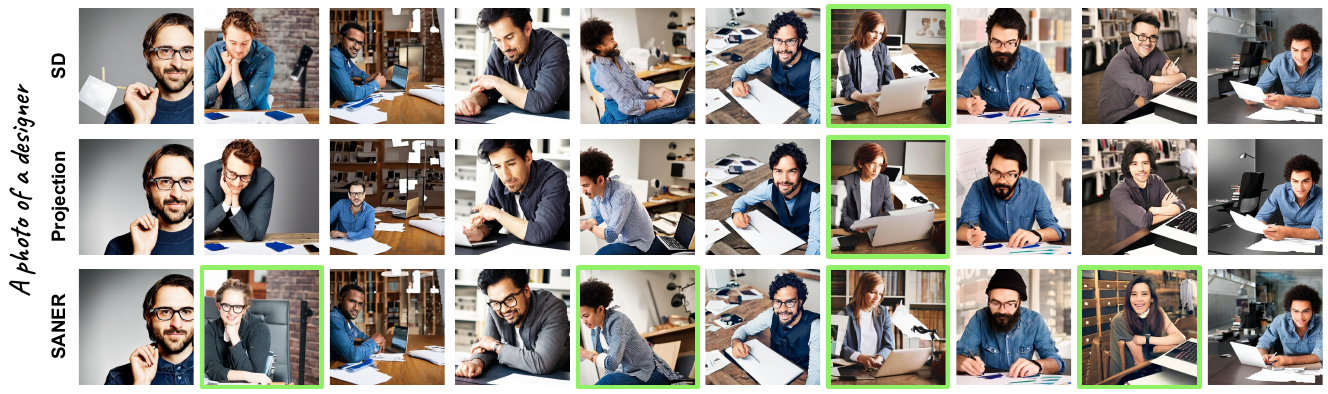}
  \vspace{-8pt}
  \caption{Generated images for the prompt, ``A photo of a designer,'' by the original Stable Diffusion (SD), projection-based debiased CLIP (Projection), and our debiased CLIP (SANER). We randomly sample $10$ images from generated images. Images framed in \textcolor{colbest}{green} denote those of the minority gender in the generated images (\ie, female).}
  \label{fig:sd-bias}
  \vspace{-11pt}
\end{figure}

We show visual examples where SANER mitigates gender bias in Fig. \ref{fig:sd-bias}. For the gender-neutral prompt, ``A photo of a designer,'' the original SD and projection-based debiasing (Projection) predominantly generate images of a man. In contrast, ours shows a more balanced gender distribution.
%

\subsection{Assessment of retention of attribute information}
\label{sec:exp-att-info}
\begin{figure}[t]
  \centering
  \includegraphics[clip, width=0.95\textwidth]{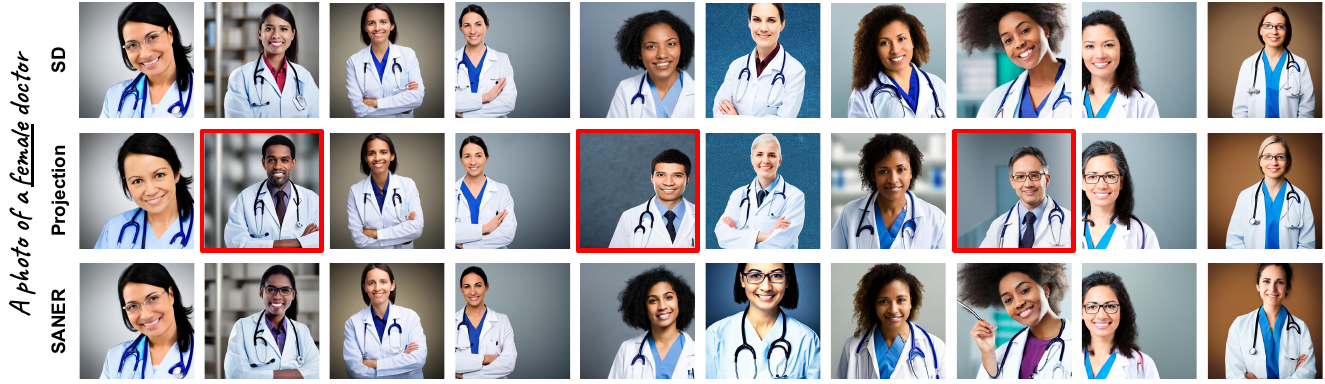}
  \vspace{-9pt}
  \caption{Generated images for the prompt, ``A photo of a female doctor,'' by the original Stable Diffusion (SD), projection-based debiased CLIP (Projection), and our debiased CLIP (SANER). \textcolor{colworst}{Red} frame indicates images with incorrect gender (\ie, male).}
  \label{fig:sd-att}
  \vspace{-15pt}
\end{figure}

Table \ref{tab:sd-gender} also shows the accuracy of how much the gender of the person in generated images matches the gender specified in the prompts. For both prompts that describe women and men, using debiased CLIP by projection-based debiasing leads to losing gender information (\ie, accuracies for projection are much lower than those for the original). Conversely, SANER retains gender information (\ie, accuracies for SANER are $1.00$). Figure \ref{fig:sd-att} confirms this, showing that using projection-based debiased CLIP results in generating male images for the prompt, ``A photo of a \underline{female} doctor''.

\vspace{-2pt}
\begin{tcolorbox}[colback=yellow!10, colframe=black, boxrule=1pt, arc=1mm]
\textbf{Insight}: Applying SANER-debiased CLIP to Stable Diffusion generates images with a more gender-equal distribution while preserving gender information in input texts.
\end{tcolorbox}

\section{Limitations}
\label{sec:limitation}


\noindent
\textbf{Further bias mitigation.}
Our experiments show that SANER noticeably reduces bias in CLIP. Nonetheless, the bias is not completely \textit{eliminated} (\eg, MaxSkew is not zero). A promising direction for further debiasing could involve debiasing the image encoder, specifically training a debiasing layer to remove attribute information from the visual features for images without human subjects.

\noindent
\textbf{Intersectional bias analysis.}
While our experiments focus on gender, age, and racial biases individually, following prior works \citep{berg2022prompt,seth2023dear,dehdashtian2024fairvlm,chuang2023debiasing}, SANER can be easily extended to various protected attributes and their combinations. For instance, considering the intersection of binary gender and age, we generate four sentences with (\textit{female}, \textit{young}), (\textit{female}, \textit{old}), (\textit{male}, \textit{young}), and (\textit{male}, \textit{old}) for the debiasing loss, \eg, ``A young woman is eating salad'' for the input text ``A woman is eating salad''. This potential for addressing complex biases is noted in future research.

\noindent
\textbf{Use of pre-defined attribute words.}
While SANER requires general lists of attributes, creating attribute lists is a one-time effort requiring minimal resources, compared to the ongoing cost and complexity of dataset annotation that often needs ethical review and domain expertise.

\section{Related Work}
\label{sec:relatedwork}


As VLMs like CLIP are applied to more tasks, concerns about their social biases have grown \citep{dehouche2021implicit,wang2021gender,ross2020measuring,wolfe2023contrastive,ruggeri2023multi,srinivasan2021worst,hirota2024resampled,hirota2024descriptive}. Hall \textit{et al.}~\citep{hall2023vision} examined gender bias in CLIP, uncovering significant discrepancies in object recognition performance based on the gender depicted in images.  such as higher accuracy in recognizing an \textit{umbrella} with \textit{women} than with \textit{men}.
These biases risk reinforcing discrimination against marginalized groups
\citep{qiu2023gender,tanjim2024discovering,wang2023tovilag}. Birhane \textit{et al.}~\citep{birhane2021multimodal}  demonstrated that adopting CLIP-based filtering in the dataset creation can select stereotypical images, like labeling a female astronaut as a "smiling housewife," leading to harmful, biased datasets.

\section{Conclusion}
\label{sec:conclusion}
This paper proposed SANER, a simple-yet-effective debiasing method for CLIP, consisting of \textbf{attribute neutralization} and \textbf{anotation-free debiasing loss}.
Consequently, SANER can leverage any image-text dataset to train the debiasing layer, outperforming existing methods in both discriminative and generative tasks (\ie, text-to-image retrieval and text-to-image generation). We also confirmed that SANER retains attribute information for attribute-specific descriptions through the gender-specified prompts for text-to-image generation. 


\bibliography{iclr2025_conference}

\begin{thebibliography}{59}
\providecommand{\natexlab}[1]{#1}
\providecommand{\url}[1]{\texttt{#1}}
\expandafter\ifx\csname urlstyle\endcsname\relax
  \providecommand{\doi}[1]{doi: #1}\else
  \providecommand{\doi}{doi: \begingroup \urlstyle{rm}\Url}\fi

\bibitem[Alabdulmohsin et~al.(2024)Alabdulmohsin, Wang, Steiner, Goyal, D'Amour, and Zhai]{alabdulmohsin2023clip}
Ibrahim Alabdulmohsin, Xiao Wang, Andreas~Peter Steiner, Priya Goyal, Alexander D'Amour, and Xiaohua Zhai.
\newblock Clip the bias: How useful is balancing data in multimodal learning?
\newblock In \emph{ICLR}, 2024.

\bibitem[Andrews et~al.(2023)Andrews, Zhao, Thong, Modas, Papakyriakopoulos, and Xiang]{andrews2023ethical}
Jerone Andrews, Dora Zhao, William Thong, Apostolos Modas, Orestis Papakyriakopoulos, and Alice Xiang.
\newblock Ethical considerations for responsible data curation.
\newblock In \emph{NeurIPS Datasets and Benchmarks Track}, 2023.

\bibitem[Bansal et~al.(2022)Bansal, Yin, Monajatipoor, and Chang]{bansal2022well}
Hritik Bansal, Da~Yin, Masoud Monajatipoor, and Kai-Wei Chang.
\newblock How well can text-to-image generative models understand ethical natural language interventions?
\newblock In \emph{EMNLP}, 2022.

\bibitem[Berg et~al.(2022)Berg, Hall, Bhalgat, Yang, Kirk, Shtedritski, and Bain]{berg2022prompt}
Hugo Berg, Siobhan~Mackenzie Hall, Yash Bhalgat, Wonsuk Yang, Hannah~Rose Kirk, Aleksandar Shtedritski, and Max Bain.
\newblock A prompt array keeps the bias away: Debiasing vision-language models with adversarial learning.
\newblock In \emph{AACL}, 2022.

\bibitem[Birhane et~al.(2021)Birhane, Prabhu, and Kahembwe]{birhane2021multimodal}
Abeba Birhane, Vinay~Uday Prabhu, and Emmanuel Kahembwe.
\newblock Multimodal datasets: Misogyny, pornography, and malignant stereotypes.
\newblock \emph{arXiv preprint arXiv:2110.01963}, 2021.

\bibitem[Burns et~al.(2018)Burns, Hendricks, Saenko, Darrell, and Rohrbach]{burns2018women}
Kaylee Burns, Lisa~Anne Hendricks, Kate Saenko, Trevor Darrell, and Anna Rohrbach.
\newblock Women also snowboard: Overcoming bias in captioning models.
\newblock In \emph{ECCV}, 2018.

\bibitem[Cho et~al.(2023)Cho, Zala, and Bansal]{cho2023dall}
Jaemin Cho, Abhay Zala, and Mohit Bansal.
\newblock Dall-eval: Probing the reasoning skills and social biases of text-to-image generation models.
\newblock In \emph{ICCV}, 2023.

\bibitem[Chuang et~al.(2023)Chuang, Jampani, Li, Torralba, and Jegelka]{chuang2023debiasing}
Ching-Yao Chuang, Varun Jampani, Yuanzhen Li, Antonio Torralba, and Stefanie Jegelka.
\newblock Debiasing vision-language models via biased prompts.
\newblock \emph{arXiv preprint arXiv:2302.00070}, 2023.

\bibitem[Das et~al.(2018)Das, Dantcheva, and Bremond]{das2018mitigating}
Abhijit Das, Antitza Dantcheva, and Francois Bremond.
\newblock Mitigating bias in gender, age and ethnicity classification: a multi-task convolution neural network approach.
\newblock In \emph{ECCV Workshops}, 2018.

\bibitem[Dehdashtian et~al.(2024)Dehdashtian, Wang, and Boddeti]{dehdashtian2024fairvlm}
Sepehr Dehdashtian, Lan Wang, and Vishnu Boddeti.
\newblock Fair{VLM}: Mitigating bias in pre-trained vision-language models.
\newblock In \emph{ICLR}, 2024.
\newblock URL \url{https://openreview.net/forum?id=HXoq9EqR9e}.

\bibitem[Dehouche(2021)]{dehouche2021implicit}
Nassim Dehouche.
\newblock Implicit stereotypes in pre-trained classifiers.
\newblock \emph{IEEE Access}, 2021.

\bibitem[Dinan et~al.(2020)Dinan, Fan, Wu, Weston, Kiela, and Williams]{dinan2020multi}
Emily Dinan, Angela Fan, Ledell Wu, Jason Weston, Douwe Kiela, and Adina Williams.
\newblock Multi-dimensional gender bias classification.
\newblock In \emph{EMNLP}, 2020.

\bibitem[Dosovitskiy et~al.(2021)Dosovitskiy, Beyer, Kolesnikov, Weissenborn, Zhai, Unterthiner, Dehghani, Minderer, Heigold, Gelly, et~al.]{dosovitskiy2020image}
Alexey Dosovitskiy, Lucas Beyer, Alexander Kolesnikov, Dirk Weissenborn, Xiaohua Zhai, Thomas Unterthiner, Mostafa Dehghani, Matthias Minderer, Georg Heigold, Sylvain Gelly, et~al.
\newblock An image is worth 16x16 words: Transformers for image recognition at scale.
\newblock In \emph{ICLR}, 2021.

\bibitem[Friedrich et~al.(2023)Friedrich, Schramowski, Brack, Struppek, Hintersdorf, Luccioni, and Kersting]{friedrich2023fair}
Felix Friedrich, Patrick Schramowski, Manuel Brack, Lukas Struppek, Dominik Hintersdorf, Sasha Luccioni, and Kristian Kersting.
\newblock Fair diffusion: Instructing text-to-image generation models on fairness.
\newblock \emph{arXiv preprint arXiv:2302.10893}, 2023.

\bibitem[Gao et~al.(2024)Gao, Geng, Zhang, Ma, Fang, Zhang, Li, and Qiao]{gao2024clip}
Peng Gao, Shijie Geng, Renrui Zhang, Teli Ma, Rongyao Fang, Yongfeng Zhang, Hongsheng Li, and Yu~Qiao.
\newblock Clip-adapter: Better vision-language models with feature adapters.
\newblock \emph{International Journal of Computer Vision}, 2024.

\bibitem[Garcia et~al.(2023)Garcia, Hirota, Wu, and Nakashima]{garcia2023uncurated}
Noa Garcia, Yusuke Hirota, Yankun Wu, and Yuta Nakashima.
\newblock Uncurated image-text datasets: Shedding light on demographic bias.
\newblock In \emph{CVPR}, 2023.

\bibitem[Geyik et~al.(2019)Geyik, Ambler, and Kenthapadi]{geyik2019fairness}
Sahin~Cem Geyik, Stuart Ambler, and Krishnaram Kenthapadi.
\newblock Fairness-aware ranking in search \& recommendation systems with application to linkedin talent search.
\newblock In \emph{SIGKDD}, 2019.

\bibitem[Gustafson et~al.(2023)Gustafson, Rolland, Ravi, Duval, Adcock, Fu, Hall, and Ross]{gustafson2023facet}
Laura Gustafson, Chloe Rolland, Nikhila Ravi, Quentin Duval, Aaron Adcock, Cheng-Yang Fu, Melissa Hall, and Candace Ross.
\newblock Facet: Fairness in computer vision evaluation benchmark.
\newblock In \emph{ICCV}, 2023.

\bibitem[Hall et~al.(2023)Hall, Gustafson, Adcock, Misra, and Ross]{hall2023vision}
Melissa Hall, Laura Gustafson, Aaron Adcock, Ishan Misra, and Candace Ross.
\newblock Vision-language models performing zero-shot tasks exhibit gender-based disparities.
\newblock In \emph{ICCV Workshops}, 2023.

\bibitem[Hausladen et~al.(2024)Hausladen, Knott, Perona, and Camerer]{hausladen2024causal}
Carina~I Hausladen, Manuel Knott, Pietro Perona, and Colin Camerer.
\newblock Causal analysis of social bias in {CLIP}, 2024.
\newblock URL \url{https://openreview.net/forum?id=Dk10QugVHb}.

\bibitem[Hessel et~al.(2021)Hessel, Holtzman, Forbes, Bras, and Choi]{hessel2021clipscore}
Jack Hessel, Ari Holtzman, Maxwell Forbes, Ronan~Le Bras, and Yejin Choi.
\newblock Clipscore: A reference-free evaluation metric for image captioning.
\newblock In \emph{EMNLP}, 2021.

\bibitem[Heusel et~al.(2017)Heusel, Ramsauer, Unterthiner, Nessler, and Hochreiter]{heusel2017gans}
Martin Heusel, Hubert Ramsauer, Thomas Unterthiner, Bernhard Nessler, and Sepp Hochreiter.
\newblock Gans trained by a two time-scale update rule converge to a local nash equilibrium.
\newblock \emph{NeurIPS}, 2017.

\bibitem[Hirota et~al.(2023)Hirota, Nakashima, and Garcia]{hirota2023model}
Yusuke Hirota, Yuta Nakashima, and Noa Garcia.
\newblock Model-agnostic gender debiased image captioning.
\newblock In \emph{CVPR}, 2023.

\bibitem[Hirota et~al.(2024{\natexlab{a}})Hirota, Andrews, Zhao, Papakyriakopoulos, Modas, Nakashima, and Xiang]{hirota2024resampled}
Yusuke Hirota, Jerone~TA Andrews, Dora Zhao, Orestis Papakyriakopoulos, Apostolos Modas, Yuta Nakashima, and Alice Xiang.
\newblock Resampled datasets are not enough: Mitigating societal bias beyond single attributes.
\newblock In \emph{EMNLP}, 2024{\natexlab{a}}.

\bibitem[Hirota et~al.(2024{\natexlab{b}})Hirota, Hachiuma, Yang, and Nakashima]{hirota2024descriptive}
Yusuke Hirota, Ryo Hachiuma, Chao-Han~Huck Yang, and Yuta Nakashima.
\newblock From descriptive richness to bias: Unveiling the dark side of generative image caption enrichment.
\newblock In \emph{EMNLP}, 2024{\natexlab{b}}.

\bibitem[Karkkainen \& Joo(2021)Karkkainen and Joo]{karkkainen2021fairface}
Kimmo Karkkainen and Jungseock Joo.
\newblock Fairface: Face attribute dataset for balanced race, gender, and age for bias measurement and mitigation.
\newblock In \emph{WACV}, 2021.

\bibitem[Kay et~al.(2017)Kay, Carreira, Simonyan, Zhang, Hillier, Vijayanarasimhan, Viola, Green, Back, Natsev, et~al.]{kay2017kinetics}
Will Kay, Joao Carreira, Karen Simonyan, Brian Zhang, Chloe Hillier, Sudheendra Vijayanarasimhan, Fabio Viola, Tim Green, Trevor Back, Paul Natsev, et~al.
\newblock The kinetics human action video dataset.
\newblock \emph{arXiv preprint arXiv:1705.06950}, 2017.

\bibitem[Krause et~al.(2013)Krause, Stark, Deng, and Fei-Fei]{krause20133d}
Jonathan Krause, Michael Stark, Jia Deng, and Li~Fei-Fei.
\newblock 3d object representations for fine-grained categorization.
\newblock In \emph{ICCV Workshops}, 2013.

\bibitem[Li et~al.(2023{\natexlab{a}})Li, Li, Savarese, and Hoi]{li2023blip}
Junnan Li, Dongxu Li, Silvio Savarese, and Steven Hoi.
\newblock Blip-2: Bootstrapping language-image pre-training with frozen image encoders and large language models.
\newblock \emph{arXiv preprint arXiv:2301.12597}, 2023{\natexlab{a}}.

\bibitem[Li et~al.(2022)Li, Zhang, Zhang, Yang, Li, Zhong, Wang, Yuan, Zhang, Hwang, Chang, and Gao]{Li2022CVPR}
Liunian~Harold Li, Pengchuan Zhang, Haotian Zhang, Jianwei Yang, Chunyuan Li, Yiwu Zhong, Lijuan Wang, Lu~Yuan, Lei Zhang, Jenq-Neng Hwang, Kai-Wei Chang, and Jianfeng Gao.
\newblock Grounded language-image pre-training.
\newblock In \emph{CVPR}, 2022.

\bibitem[Li et~al.(2023{\natexlab{b}})Li, Zhu, Wen, and Yang]{li2023decap}
Wei Li, Linchao Zhu, Longyin Wen, and Yi~Yang.
\newblock Decap: Decoding clip latents for zero-shot captioning via text-only training.
\newblock In \emph{ICLR}, 2023{\natexlab{b}}.

\bibitem[Lin et~al.(2014)Lin, Maire, Belongie, Hays, Perona, Ramanan, Doll{\'a}r, and Zitnick]{lin2014microsoft}
Tsung-Yi Lin, Michael Maire, Serge Belongie, James Hays, Pietro Perona, Deva Ramanan, Piotr Doll{\'a}r, and C~Lawrence Zitnick.
\newblock Microsoft {COCO}: Common objects in context.
\newblock In \emph{ECCV}, 2014.

\bibitem[Liu et~al.(2024{\natexlab{a}})Liu, Li, Wu, and Lee]{liu2024visual}
Haotian Liu, Chunyuan Li, Qingyang Wu, and Yong~Jae Lee.
\newblock Visual instruction tuning.
\newblock \emph{NeurIPS}, 2024{\natexlab{a}}.

\bibitem[Liu et~al.(2024{\natexlab{b}})Liu, Schaldenbrand, Okogwu, Peng, Yun, Hundt, Kim, and Oh]{liu2024scoft}
Zhixuan Liu, Peter Schaldenbrand, Beverley-Claire Okogwu, Wenxuan Peng, Youngsik Yun, Andrew Hundt, Jihie Kim, and Jean Oh.
\newblock Scoft: Self-contrastive fine-tuning for equitable image generation.
\newblock In \emph{CVPR}, 2024{\natexlab{b}}.

\bibitem[L\"uddecke \& Ecker(2022)L\"uddecke and Ecker]{Luddecke2022CVPR}
Timo L\"uddecke and Alexander Ecker.
\newblock Image segmentation using text and image prompts.
\newblock In \emph{CVPR}, 2022.

\bibitem[Mokady et~al.(2021)Mokady, Hertz, and Bermano]{mokady2021clipcap}
Ron Mokady, Amir Hertz, and Amit~H Bermano.
\newblock Clipcap: Clip prefix for image captioning.
\newblock \emph{arXiv preprint arXiv:2111.09734}, 2021.

\bibitem[Nair \& Hinton(2010)Nair and Hinton]{nair2010rectified}
Vinod Nair and Geoffrey~E Hinton.
\newblock Rectified linear units improve restricted boltzmann machines.
\newblock In \emph{ICML}, 2010.

\bibitem[Qiu et~al.(2023)Qiu, Dou, Wang, Celikyilmaz, and Peng]{qiu2023gender}
Haoyi Qiu, Zi-Yi Dou, Tianlu Wang, Asli Celikyilmaz, and Nanyun Peng.
\newblock Gender biases in automatic evaluation metrics for image captioning.
\newblock In \emph{EMNLP}, 2023.

\bibitem[Radford et~al.(2021)Radford, Kim, Hallacy, Ramesh, Goh, Agarwal, Sastry, Askell, Mishkin, Clark, et~al.]{radford2021learning}
Alec Radford, Jong~Wook Kim, Chris Hallacy, Aditya Ramesh, Gabriel Goh, Sandhini Agarwal, Girish Sastry, Amanda Askell, Pamela Mishkin, Jack Clark, et~al.
\newblock Learning transferable visual models from natural language supervision.
\newblock In \emph{ICML}, 2021.

\bibitem[Ramaswamy et~al.(2021)Ramaswamy, Kim, and Russakovsky]{ramaswamy2020debiasing}
Vikram~V. Ramaswamy, Sunnie S.~Y. Kim, and Olga Russakovsky.
\newblock Fair attribute classification through latent space de-biasing.
\newblock In \emph{IEEE/CVF Conference on Computer Vision and Pattern Recognition (CVPR)}, 2021.

\bibitem[Rombach et~al.(2022)Rombach, Blattmann, Lorenz, Esser, and Ommer]{rombach2022high}
Robin Rombach, Andreas Blattmann, Dominik Lorenz, Patrick Esser, and Bj{\"o}rn Ommer.
\newblock High-resolution image synthesis with latent diffusion models.
\newblock In \emph{CVPR}, 2022.

\bibitem[Ross et~al.(2021)Ross, Katz, and Barbu]{ross2020measuring}
Candace Ross, Boris Katz, and Andrei Barbu.
\newblock Measuring social biases in grounded vision and language embeddings.
\newblock In \emph{ACL}, 2021.

\bibitem[Ruggeri et~al.(2023)Ruggeri, Nozza, et~al.]{ruggeri2023multi}
Gabriele Ruggeri, Debora Nozza, et~al.
\newblock A multi-dimensional study on bias in vision-language models.
\newblock In \emph{Findings of ACL}, 2023.

\bibitem[Russakovsky et~al.(2015)Russakovsky, Deng, Su, Krause, Satheesh, Ma, Huang, Karpathy, Khosla, Bernstein, Berg, and Fei-Fei]{imagenet}
Olga Russakovsky, Jia Deng, Hao Su, Jonathan Krause, Sanjeev Satheesh, Sean Ma, Zhiheng Huang, Andrej Karpathy, Aditya Khosla, Michael Bernstein, Alexander~C. Berg, and Li~Fei-Fei.
\newblock {ImageNet Large Scale Visual Recognition Challenge}.
\newblock \emph{IJCV}, 2015.

\bibitem[Seth et~al.(2023)Seth, Hemani, and Agarwal]{seth2023dear}
Ashish Seth, Mayur Hemani, and Chirag Agarwal.
\newblock Dear: Debiasing vision-language models with additive residuals.
\newblock In \emph{CVPR}, 2023.

\bibitem[Shen et~al.(2022)Shen, Li, Tan, Bansal, Rohrbach, Chang, Yao, and Keutzer]{shen2021much}
Sheng Shen, Liunian~Harold Li, Hao Tan, Mohit Bansal, Anna Rohrbach, Kai-Wei Chang, Zhewei Yao, and Kurt Keutzer.
\newblock How much can clip benefit vision-and-language tasks?
\newblock In \emph{ICLR}, 2022.

\bibitem[Srinivasan \& Bisk(2022)Srinivasan and Bisk]{srinivasan2021worst}
Tejas Srinivasan and Yonatan Bisk.
\newblock Worst of both worlds: Biases compound in pre-trained vision-and-language models.
\newblock In \emph{ACL Workshops}, 2022.

\bibitem[Tanjim et~al.(2024)Tanjim, Singh, Kafle, Sinha, and Cottrell]{tanjim2024discovering}
Md~Mehrab Tanjim, Krishna~Kumar Singh, Kushal Kafle, Ritwik Sinha, and Garrison~W Cottrell.
\newblock Discovering and mitigating biases in clip-based image editing.
\newblock In \emph{WACV}, 2024.

\bibitem[Tao et~al.(2023)Tao, Bao, Tang, and Xu]{tao2023galip}
Ming Tao, Bing-Kun Bao, Hao Tang, and Changsheng Xu.
\newblock Galip: Generative adversarial clips for text-to-image synthesis.
\newblock In \emph{CVPR}, 2023.

\bibitem[Teo et~al.(2023)Teo, Abdollahzadeh, and Cheung]{teo2024measuring}
Christopher Teo, Milad Abdollahzadeh, and Ngai-Man~Man Cheung.
\newblock On measuring fairness in generative models.
\newblock In \emph{NeurIPS}, 2023.

\bibitem[Tewel et~al.(2022)Tewel, Shalev, Schwartz, and Wolf]{tewel2022zerocap}
Yoad Tewel, Yoav Shalev, Idan Schwartz, and Lior Wolf.
\newblock Zerocap: Zero-shot image-to-text generation for visual-semantic arithmetic.
\newblock In \emph{CVPR}, 2022.

\bibitem[Wang et~al.(2021)Wang, Liu, and Wang]{wang2021gender}
Jialu Wang, Yang Liu, and Xin~Eric Wang.
\newblock Are gender-neutral queries really gender-neutral? mitigating gender bias in image search.
\newblock \emph{arXiv preprint arXiv:2109.05433}, 2021.

\bibitem[Wang et~al.(2023)Wang, Yi, Jiang, Zhou, Wei, and Xie]{wang2023tovilag}
Xinpeng Wang, Xiaoyuan Yi, Han Jiang, Shanlin Zhou, Zhihua Wei, and Xing Xie.
\newblock Tovilag: Your visual-language generative model is also an evildoer.
\newblock In \emph{EMNLP}, 2023.

\bibitem[Wolfe \& Caliskan(2022)Wolfe and Caliskan]{wolfe2022markedness}
Robert Wolfe and Aylin Caliskan.
\newblock Markedness in visual semantic ai.
\newblock In \emph{FAccT}, 2022.

\bibitem[Wolfe et~al.(2023)Wolfe, Yang, Howe, and Caliskan]{wolfe2023contrastive}
Robert Wolfe, Yiwei Yang, Bill Howe, and Aylin Caliskan.
\newblock Contrastive language-vision ai models pretrained on web-scraped multimodal data exhibit sexual objectification bias.
\newblock In \emph{FAccT}, 2023.

\bibitem[Yamazaki et~al.(2022)Yamazaki, Truong, Vo, Kidd, Rainwater, Luu, and Le]{yamazaki2022vlcap}
Kashu Yamazaki, Sang Truong, Khoa Vo, Michael Kidd, Chase Rainwater, Khoa Luu, and Ngan Le.
\newblock Vlcap: Vision-language with contrastive learning for coherent video paragraph captioning.
\newblock In \emph{ICIP}, 2022.

\bibitem[Yamazaki et~al.(2023)Yamazaki, Vo, Truong, Raj, and Le]{yamazaki2023vltint}
Kashu Yamazaki, Khoa Vo, Quang~Sang Truong, Bhiksha Raj, and Ngan Le.
\newblock Vltint: visual-linguistic transformer-in-transformer for coherent video paragraph captioning.
\newblock In \emph{AAAI}, 2023.

\bibitem[Zhao et~al.(2017)Zhao, Wang, Yatskar, Ordonez, and Chang]{zhao2017mals}
Jieyu Zhao, Tianlu Wang, Mark Yatskar, Vicente Ordonez, and Kai-Wei Chang.
\newblock Men also like shopping: Reducing gender bias amplification using corpus-level constraints.
\newblock In \emph{EMNLP}, 2017.

\bibitem[Zhong et~al.(2022)Zhong, Yang, Zhang, Li, Codella, Li, Zhou, Dai, Yuan, Li, and Gao]{Zhong2022CVPR}
Yiwu Zhong, Jianwei Yang, Pengchuan Zhang, Chunyuan Li, Noel Codella, Liunian~Harold Li, Luowei Zhou, Xiyang Dai, Lu~Yuan, Yin Li, and Jianfeng Gao.
\newblock Regionclip: Region-based language-image pretraining.
\newblock In \emph{CVPR}, 2022.

\end{thebibliography}
\bibliographystyle{iclr2025_conference}

\appendix

\section*{Appendix}
This appendix includes:

\begin{itemize}
    
    \item Implementation details for SANER (Appendix~\ref{sec:details}).
    \item Further analysis (Appendix~\ref{sec:further}).
    \item List of person-, gender-, age-, and race-specific terms (Appendix~\ref{sec:person-terms}).
    \item List of concepts and occupations (Appendix~\ref{sec:concepts}).
    \item Additional visual examples for image generation experiments (Appendix~\ref{sec:examples}).    
    \item Potential extension (Appendix~\ref{sec:extension}).
    \item Potential negative impact (Appendix~\ref{sec:impact}).
\end{itemize}

\section{Implementation Details for SANER}
\label{sec:details}
We train the debiasing layer (a multilayer perception with two linear layers with ReLU activation) using $170,624$ image-caption pairs, which is a subset of the COCO training set with person-related words/phrases defined in Sec. \ref{sec:person-terms}. The hidden embedding dimensionality for the debiasing layer is set to $128$. We empirically set $\alpha$, $\beta$, and $\gamma$ to $1.0$, $0.1$, and $0.0001$ (Eq. (11) in the main paper), respectively. We set the training epochs, batch size, and learning rate to $5$, $128$, and $5\times10^{-6}$, respectively. The training is conducted with a machine equipped with a single NVIDIA A100 GPU 40GB, and it took five hours to train the debiasing layer. Note that the weights of the debiasing layer are updated, and the weights for the rest of the modules are frozen.

\textbf{Attribute-specific description generation. }To avoid computational complexity, we implement a systematic approach using predefined mapping dictionaries between gender terms (\eg, ``woman'' $\rightarrow$ ``man'', ``she'' $\rightarrow$ ``he''). Our algorithm carefully identifies and replaces gender-specific tokens while preserving sentence structure, ensuring no ungrammatical combinations (like ``A he/she'') are generated. This ensures efficient and coherent text generation that maintains natural language patterns.

\textbf{Racial bias mitigation.} 
For race attribute, we remove race-specific terms\footnote{We define race-specific terms. The list is in Section \ref{sec:person-terms}.} (\eg, \textit{African} and \textit{Asian}) in text descriptions, for instance, ``An African woman is eating salad'' $\rightarrow$ ``A woman is eating salad''. 

\section{Additional Experiments}
\label{sec:further}

\subsection{Complete results for age and racial biases}
In Table \ref{tab:app-age-retrieve} and \ref{tab:app-race-retrieve}, we show the complete results of Table \ref{tab:age-retrieve}, including results on PATA. The results further verify that SANER demonstrates superior performance in mitigating age and racial biases compared to the existing method.

\subsection{Loss ablation}
To validate the effectiveness of each regularization loss (\ie, reconstruction loss and contrastive loss), we conduct an ablation study by removing one of the losses or both losses. Table \ref{tab:loss} presents the results of gender bias.
The results show that using both reconstruction and contrastive losses yields the best results regarding gender bias mitigation and zero-shot classification accuracy on ImangeNet \citep{imagenet}. Furthermore, SANER without regularization losses significantly degrades CLIP's zero-shot classification ability (\ie, from $65.4$ to $58.8$). These observations confirm the importance of having both reconstruction and contrastive losses for the regularization. 

\textbf{Negative impact on adding only the contrastive loss.} When using only the contrastive loss, MaxSkew scores on FairFace increase compared to no regularization (18.0/19.4/21.7 vs. 15.7/15.0/15.3 without regularization), indicating less effective debiasing. This occurs because the contrastive loss alone focuses on maintaining image-text alignment but does not constrain the debiased features to remain close to the original CLIP features. As a result, the feature modifications may become suboptimal for bias removal. The reconstruction loss plays a crucial role in ensuring the modified features preserve essential semantic information while effectively removing unwanted bias.

\begin{table}[t]
\renewcommand{\arraystretch}{1.1}
\setlength{\tabcolsep}{5pt}
\scriptsize
\centering
\caption{\textbf{Age bias}, evaluated by MaxSkew@1000 (scaled by $100$), on FairFace and PATA. \textbf{Bold} denotes the best across the models.}
\vspace{-5pt}
\begin{tabularx}{0.835\columnwidth}{l r r r r r r r}
\toprule
 & \multicolumn{3}{c}{FairFace} &&\multicolumn{3}{c}{PATA}\\ 
\cline{2-4} 
\cline{6-8}
\multirow{-2}{*}{CLIP Model} & \multirow{1.3}{*}{Adjective} & \multirow{1.3}{*}{Occupation} & \multirow{1.3}{*}{Activity} & & \multirow{1.3}{*}{Adjective} & \multirow{1.3}{*}{Occupation} & \multirow{1.3}{*}{Activity}   \\
\midrule
Original\citep{radford2021learning}  & $111.1$ & $121.1$ & $113.0$ && $40.4$ & $44.4$ & $39.7$ \\
\midrule
Projection \citep{chuang2023debiasing} & $107.6$ & $112.8$ & $\mathbf{100.0}$ && $37.6$ & $45.6$ & $45.5$ \\
\cellcolor{oursrow}SANER (Ours) & \cellcolor{oursrow}$\mathbf{96.0}$ & \cellcolor{oursrow}$\mathbf{112.6}$ & \cellcolor{oursrow}$101.9$ &\cellcolor{oursrow}& \cellcolor{oursrow}$\mathbf{30.3}$ & \cellcolor{oursrow}$\mathbf{36.7}$ & \cellcolor{oursrow}$\mathbf{27.5}$ \\
\bottomrule
\end{tabularx}
\vspace{-5pt}
\label{tab:app-age-retrieve}
\end{table}

\begin{table}[t]
\renewcommand{\arraystretch}{1.1}
\setlength{\tabcolsep}{5pt}
\scriptsize
\centering
\caption{\textbf{Racial bias}, evaluated by MaxSkew@1000 (scaled by $100$), on FairFace and PATA. \textbf{Bold} denotes the best across the models.}
\vspace{-5pt}
\begin{tabularx}{0.835\columnwidth}{l r r r r r r r}
\toprule
 & \multicolumn{3}{c}{FairFace} &&\multicolumn{3}{c}{PATA}\\ 
\cline{2-4} 
\cline{6-8}
\multirow{-2}{*}{CLIP Model} & \multirow{1.3}{*}{Adjective} & \multirow{1.3}{*}{Occupation} & \multirow{1.3}{*}{Activity} & & \multirow{1.3}{*}{Adjective} & \multirow{1.3}{*}{Occupation} & \multirow{1.3}{*}{Activity}   \\
\midrule
Original\citep{radford2021learning}  & $62.2$ & $57.4$ & $68.3$ && $33.4$ & $28.6$ & $31.5$ \\
\midrule
Projection \citep{chuang2023debiasing} & $56.9$ & $75.3$ & $67.0$ && $\mathbf{19.9}$ & $43.8$ & $26.3$ \\
\cellcolor{oursrow}SANER (Ours) & \cellcolor{oursrow}$\mathbf{49.3}$ & \cellcolor{oursrow}$\mathbf{45.7}$ & \cellcolor{oursrow}$\mathbf{46.6}$ &\cellcolor{oursrow}& \cellcolor{oursrow}$28.9$ & \cellcolor{oursrow}$\mathbf{21.2}$ & \cellcolor{oursrow}$\mathbf{20.5}$ \\
\bottomrule
\end{tabularx}
\vspace{-5pt}
\label{tab:app-race-retrieve}
\end{table}

\begin{table}[t]
\renewcommand{\arraystretch}{1.1}
\setlength{\tabcolsep}{5pt}
\scriptsize
\centering
\caption{Gender bias, evaluated by MaxSkew@1000 (scaled by $100$), on FairFace and PATA for our method (SANER) with different regularization loss combinations. Recon denotes the use of the reconstruction loss, and cont represents the use of the contrastive loss. IN acc is the zero-shot classification accuracy on ImageNet. Adj, Occ, and Act represent the types of concepts (\ie, Adjective, Occupations, and Activity, respectively). A lower value is better (less gender bias). \textbf{Bold} represents the best across the SANER variants.}
\vspace{-5pt}
\begin{tabularx}{0.76\columnwidth}{l r r r r r r r r r}
\toprule
& \multicolumn{3}{c}{FairFace} &&\multicolumn{3}{c}{PATA}\\ 
\cline{2-4} 
\cline{6-8}
\multirow{-2}{*}{CLIP Model} & \multirow{1.3}{*}{Adj} & \multirow{1.3}{*}{Occ} & \multirow{1.3}{*}{Act} & & \multirow{1.3}{*}{Adj} & \multirow{1.3}{*}{Occ} & \multirow{1.3}{*}{Act} &&  IN acc  (\%) \\
\midrule
Original \citep{radford2021learning}  & $22.9$ & $33.7$ & $19.5$ && $12.1$ & $18.7$ & $10.7$ && $65.4$\\
\midrule
\cellcolor{oursrow}SANER & \cellcolor{oursrow} & \cellcolor{oursrow}& \cellcolor{oursrow}& \cellcolor{oursrow}& \cellcolor{oursrow}& \cellcolor{oursrow}& \cellcolor{oursrow} &\cellcolor{oursrow}&\cellcolor{oursrow}\\
(no regularization) & $15.7$ & $15.0$ & $15.3$ && $15.8$ & $14.4$ & $15.4$ &&$58.8$ \\
Recon & $10.2$ & $16.7$ & $11.4$ && $6.2$ & $12.0$ & $6.2$ &&$64.2$ \\
Cont & $18.0$ & $19.4$ & $21.7$ && $7.0$ & $\mathbf{9.1}$ & $9.0$ &&$63.0$ \\
Recon+Cont & $\mathbf{8.9}$ & $\mathbf{14.5}$ & $\mathbf{7.7}$ && $\mathbf{5.4}$ & $9.5$ & $\mathbf{3.3}$ && $\mathbf{65.2}$\\
\bottomrule
\end{tabularx}
\label{tab:loss}
\end{table}

\subsection{Analysis on the data size}

The experiments in the main paper (Sections 5 and 6) verify that SANER outperforms existing debiasing methods, showing a better bias mitigation ability in terms of gender and age biases. This superior performance of SANER may be because SANER is trained with diverse text descriptions (\ie, captions in COCO \citep{lin2014microsoft}), which are not constrained like pre-defined concepts required in the previous method \citep{berg2022prompt}. In this section, we conduct an experiment to verify this hypothesis. Specifically, we use $n$ percent of the training samples (\ie, $17,624$ image-caption pairs of COCO) to evaluate the impact of the training dataset size. We use the same settings in Sec. \ref{sec:details}, but use the different training epochs to align the number of iterations. The results are shown in Table \ref{tab:size}. 

The results validate that as the number of data samples increases, gender bias is reduced. Specifically, while using a part of the training data results in mitigating gender bias (\ie, MaxSkew scores are smaller than the original CLIP), using the full training samples (\ie, COCO-100\%) gives the best results, showing the importance of the use of more diverse data for debiasing. 

\begin{table}[t]
\renewcommand{\arraystretch}{1.1}
\setlength{\tabcolsep}{5pt}
\scriptsize
\centering
\caption{Gender bias, evaluated by MaxSkew@1000 (scaled by $100$), on FairFace and PATA for the original CLIP (Original) and our method (SANER) with different data sizes. COCO-$n$\% denotes that we use $n$\% of the training samples. IN acc is the zero-shot classification accuracy on ImageNet. A lower value is better (less gender bias). \textbf{Bold} represents the best across the SANER variants.}
\vspace{-5pt}
\begin{tabularx}{0.76\columnwidth}{l r r r r r r r r r}
\toprule
& \multicolumn{3}{c}{FairFace} &&\multicolumn{3}{c}{PATA}\\ 
\cline{2-4} 
\cline{6-8}
\multirow{-2}{*}{CLIP Model} & \multirow{1.3}{*}{Adj} & \multirow{1.3}{*}{Occ} & \multirow{1.3}{*}{Act} & & \multirow{1.3}{*}{Adj} & \multirow{1.3}{*}{Occ} & \multirow{1.3}{*}{Act} && IN acc (\%) \\
\midrule
Original \citep{radford2021learning}  & $22.9$ & $33.7$ & $19.5$ && $12.1$ & $18.7$ & $10.7$ && $65.4$\\
\midrule
\cellcolor{oursrow}SANER & \cellcolor{oursrow} & \cellcolor{oursrow}& \cellcolor{oursrow}& \cellcolor{oursrow}& \cellcolor{oursrow}& \cellcolor{oursrow}& \cellcolor{oursrow} & \cellcolor{oursrow} & \cellcolor{oursrow} \\
COCO-50\%  & $15.7$ & $19.4$ & $17.9$ && $10.0$ & $14.1$ & $10.2$ && $\mathbf{65.4}$\\
COCO-75\%  & $12.8$ & $17.6$ & $16.1$ && $7.8$ & $13.0$ & $8.3$ && $\mathbf{65.4}$\\
COCO-100\%  & $\mathbf{8.9}$ & $\mathbf{14.5}$ & $\mathbf{7.7}$ && $\mathbf{5.4}$ & $\mathbf{9.5}$ & $\mathbf{3.3}$ && $65.2$\\
\bottomrule
\end{tabularx}
\label{tab:size}
\end{table}

\subsection{Quality of the generated images}

We evaluate image fidelity and image-text alignment on the COCO Karpathy test set. Specifically, we use FID score \citep{heusel2017gans} and CLIPScore \citep{hessel2021clipscore} to measure fidelity and image-text alignment, respectively. The results show that SANER achieves comparable performance to the original Stable Diffusion (FID: 28.1, CLIPScore: 31.2 vs. FID: 28.7, CLIPScore: 31.0 for SANER), confirming that SANER achieves debiasing while maintaining image generation quality. 

\subsection{Experiments on BLIP}

To verify the effectiveness of SANER for VLMs beyond CLIP, we conduct gender bias experiments using BLIP \citep{li2023blip}. As shown in Table \ref{tab:blip}, SANER demonstrates superior debiasing performance compared to the original BLIP and the projection-based debiasing.

\begin{table}[h]
\renewcommand{\arraystretch}{1.1}
\setlength{\tabcolsep}{5pt}
\scriptsize
\centering
\caption{Gender bias for \textbf{BLIP}, evaluated by MaxSkew@1000.}
\vspace{-5pt}
\begin{tabularx}{0.72\columnwidth}{l r r r r r r r}
\toprule
& \multicolumn{3}{c}{FairFace} &&\multicolumn{3}{c}{PATA}\\ 
\cline{2-4} 
\cline{6-8}
\multirow{-2}{*}{BLIP Model} & \multirow{1.3}{*}{Adjective} & \multirow{1.3}{*}{Occupation} & \multirow{1.3}{*}{Activity} & & \multirow{1.3}{*}{Adjective} & \multirow{1.3}{*}{Occupation} & \multirow{1.3}{*}{Activity}   \\
\midrule
Original   & $16.8$ & $15.3$ & $12.8$ && $7.7$ & $7.4$ & $5.5$ \\
\midrule
Projection & $\mathbf{14.9}$ & $19.8$ & $17.4$ && $12.0$ & $14.7$ & $17.4$ \\
\cellcolor{oursrow}SANER (Ours)  & \cellcolor{oursrow}$15.3$ & \cellcolor{oursrow}$\mathbf{13.9}$ & \cellcolor{oursrow}$\mathbf{10.9}$ &\cellcolor{oursrow}& \cellcolor{oursrow}$\mathbf{6.7}$ & \cellcolor{oursrow}$\mathbf{6.7}$ & \cellcolor{oursrow}$\mathbf{4.9}$ \\
\bottomrule
\end{tabularx}
\label{tab:blip}
\end{table}

\subsection{Experiments on FACET}

In addition to FairFace and PATA, we evaluate SANER on the FACET dataset \citep{gustafson2023facet}. Similar to PATA, FACET comprises real-world, natural images but includes a broader range of annotations for protected attributes, such as hair color. To evaluate SANER's effectiveness on FACET, which differs in data distribution from FairFace and PATA, we conducted experiments focusing on gender bias. The results, presented in Table~\ref{tab:facet}, demonstrate SANER's superior debiasing performance, further supporting its robustness across diverse dataset distributions. Experiments on other attributes, such as hair color, are left for future work.

\begin{table}[h]
\renewcommand{\arraystretch}{1.1}
\setlength{\tabcolsep}{5pt}
\small
\centering
\caption{Gender bias on \textbf{FACET}, evaluated by MaxSkew@1000.}
\vspace{-5pt}
\begin{tabularx}{0.51\columnwidth}{l r r r}
\toprule
& \multicolumn{3}{c}{FACET} \\ 
\cline{2-4} 
\multirow{-2}{*}{CLIP Model} & \multirow{1.3}{*}{Adjective} & \multirow{1.3}{*}{Occupation} & \multirow{1.3}{*}{Activity}  \\
\midrule
Original   & $46.0$ & $37.7$ & $33.4$ \\
\midrule
Projection & $37.3$ & $46.3$ & $41.1$  \\
\cellcolor{oursrow}SANER (Ours)  & \cellcolor{oursrow}$\mathbf{31.7}$ & \cellcolor{oursrow}$\mathbf{24.7}$ & \cellcolor{oursrow}$\mathbf{25.2}$  \\
\bottomrule
\end{tabularx}
\label{tab:facet}
\end{table}

\section{List of Person-, Gender-, Age-, Race-Specific Terms}
\label{sec:person-terms}
The \textbf{person-related words} that are used to identify text descriptions that are relevant to humans (in Sec. 4.1 in the main paper) are as below:

\noindent
\textit{actor}, \textit{actress}, \textit{adult}, \textit{architect}, \textit{artist}, \textit{associate}, \textit{aunt}, \textit{baby}, \textit{boy}, \textit{boyfriend}, \textit{brother}, \textit{chairman}, \textit{chairperson}, \textit{chairwoman}, \textit{chef}, \textit{child}, \textit{coach}, \textit{colleague}, \textit{comedian}, \textit{counselor}, \textit{cowboy}, \textit{cowgirl}, \textit{dancer}, \textit{daughter}, \textit{designer}, \textit{director}, \textit{doctor}, \textit{driver}, \textit{dude}, \textit{elder}, \textit{emperor}, \textit{employee}, \textit{employer}, \textit{engineer}, \textit{entrepreneur}, \textit{executive}, \textit{expecting}, \textit{father}, \textit{female}, \textit{friend}, \textit{gentleman}, \textit{girl}, \textit{girlfriend}, \textit{guy}, \textit{hairdresser}, \textit{he}, \textit{her}, \textit{hers}, \textit{herself}, \textit{him}, \textit{himself}, \textit{his}, \textit{husband}, \textit{individual}, \textit{infant}, \textit{instructor}, \textit{kid}, \textit{lady}, \textit{lawyer}, \textit{leader}, \textit{lecturer}, \textit{male}, \textit{man}, \textit{manager}, \textit{mechanic}, \textit{member}, \textit{mentor}, \textit{mother}, \textit{musician}, \textit{neighbor}, \textit{novelist}, \textit{nurse}, \textit{parent}, \textit{partner}, \textit{people}, \textit{performer}, \textit{person}, \textit{pharmacist}, \textit{photographer}, \textit{physician}, \textit{pilot}, \textit{player}, \textit{police officer}, \textit{policeman}, \textit{policewoman}, \textit{politician}, \textit{pregnant}, \textit{prince}, \textit{princess}, \textit{professor}, \textit{queen}, \textit{relative}, \textit{researcher}, \textit{royal}, \textit{scholar}, \textit{scientist}, \textit{secretary}, \textit{server}, \textit{she}, \textit{sibling}, \textit{singer}, \textit{sister}, \textit{son}, \textit{specialist}, \textit{spouse}, \textit{student}, \textit{surfer}, \textit{surgeon}, \textit{tailor}, \textit{teacher}, \textit{technician}, \textit{teenager}, \textit{their}, \textit{theirs}, \textit{them}, \textit{themselves}, \textit{therapist}, \textit{they}, \textit{toddler}, \textit{uncle}, \textit{veterinarian}, \textit{volunteer}, \textit{waiter}, \textit{waitress}, \textit{wife}, \textit{woman}, \textit{worker}, \textit{writer}, \textit{youth}, and their plurals.

We list the \textbf{gender-specific terms} that are used to create attribute-neutral text descriptions $\xi_n(t)$ (in Sec. 4.1 in the main paper):  
\textcolor{orange}{\textit{woman}}, \textcolor{orange}{\textit{female}}, \textcolor{orange}{\textit{lady}}, \textcolor{orange}{\textit{mother}}, \textcolor{orange}{\textit{girl}}, \textcolor{orange}{\textit{aunt}}, 
\textcolor{orange}{\textit{wife}}, \textcolor{orange}{\textit{actress}}, \textcolor{orange}{\textit{princess}}, \textcolor{orange}{\textit{waitress}}, \textcolor{orange}{\textit{sister}}, \textcolor{orange}{\textit{queen}}, \textcolor{orange}{\textit{pregnant}}, \textcolor{orange}{\textit{daughter}}, \textcolor{orange}{\textit{she}}, \textcolor{orange}{\textit{her}}, 
\textcolor{orange}{\textit{hers}}, \textcolor{orange}{\textit{herself}}, \textcolor{olive}{\textit{man}}, \textcolor{olive}{\textit{male}}, \textcolor{olive}{\textit{father}}, \textcolor{olive}{\textit{gentleman}}, \textcolor{olive}{\textit{boy}}, \textcolor{olive}{\textit{uncle}}, \textcolor{olive}{\textit{husband}}, \textcolor{olive}{\textit{actor}}, \textcolor{olive}{\textit{prince}}, \textcolor{olive}{\textit{waiter}}, \textcolor{olive}{\textit{son}}, \textcolor{olive}{\textit{brother}}, \textcolor{olive}{\textit{guy}}, \textcolor{olive}{\textit{emperor}}, \textcolor{olive}{\textit{dude}}, \textcolor{olive}{\textit{cowboy}}, \textcolor{olive}{\textit{he}}, \textcolor{olive}{\textit{his}}, \textcolor{olive}{\textit{him}}, \textcolor{olive}{\textit{himself}} and their plurals (\textcolor{orange}{orange} denotes female-specific words, and \textcolor{olive}{olive} represents male-specific terms). To synthesize attribute-specific descriptions (\ie, $\mathcal{T} = \{\xi_g(t)|t \in \mathcal{D}, g \in \mathcal{A}\}$ in Sec. 4.3 in the main paper) for binary gender, we replace person-specific terms in the attribute-neutral descriptions with their corresponding gender terms (\eg, \textit{person} $\rightarrow$ \textit{woman} and \textit{person} $\rightarrow$ \textit{man}).

We also list the \textbf{age-specific terms} used to create attribute-neutral text descriptions $\xi_n(t)$ (in Sec. 4.1 in the main paper): 
\textit{elderly} ,\textit{baby}, \textit{child}, \textit{kid}, \textit{teenager}, \textit{adult}, \textit{youth}, \textit{infant}, \textit{toddler}, \textit{elder}, \textit{girl}, \textit{boy}, \textit{young}, \textit{old}, \textit{teenage}, and their plurals. To create attribute-specific descriptions (\ie, $\mathcal{T} = \{\xi_g(t)|t \in \mathcal{D}, g \in \mathcal{A}\}$ in Sec. 4.3 in the main paper) for binary age, we add \textit{young} or \textit{old} just before the person-specific terms (\eg, \textit{person} $\rightarrow$ \textit{young person} and \textit{person} $\rightarrow$ \textit{old person}).

Regarding race attributes, we use \textbf{race-specific} terms to create attribute-neutral text descriptions: \textit{african}, \textit{africa}, \textit{asian}, \textit{oriental}, \textit{asia}, \textit{east asian}, \textit{south asian}, \textit{south east asian}, \textit{black}, \textit{caucasian}, \textit{european}, \textit{hispanic}, \textit{latino}, \textit{latina}, \textit{latinx}, \textit{white}, \textit{arab},
\textit{arabic}, \textit{middle eastern}, \textit{native}, \textit{indigenous}, \textit{american}, \textit{african american}, \textit{usa}, \textit{united states}, \textit{chinese}, \textit{china}, \textit{japanese}, \textit{japan}, \textit{indian}, \textit{india}, \textit{mexican}, \textit{mexico}, \textit{italian}, \textit{italy}, \textit{spanish}, \textit{german}, \textit{french}, \textit{france},
\textit{english}, \textit{british}, \textit{england}, \textit{russian}, \textit{swiss}, \textit{hawaiian}, \textit{thai}, \textit{brazil}, \textit{brazilian}, \textit{canadian}, \textit{canada}, \textit{australian}, \textit{australia}, \textit{new zealander}, \textit{new zealand}, \textit{korean}, \textit{korea},
\textit{filipino}, \textit{philippines}, \textit{vietnamese}, \textit{vietnam}, \textit{malaysian}, \textit{malaysia}, \textit{singaporean}, \textit{singapore}, \textit{indonesian}, \textit{indonesia}, \textit{thai}, \textit{thailand}, \textit{burmese}, \textit{myanmar}, \textit{cambodian}, \textit{cambodia},
\textit{laotian}, \textit{laos}, \textit{taiwanese}, \textit{taiwan}, \textit{pacific}, \textit{melanesian}, \textit{melanesia}, \textit{polynesian}, \textit{polynesia}, \textit{micronesian}, \textit{micronesia}, \textit{aboriginal}, \textit{aborigine}. We adopt racial classes used in FairFace (\ie, East Asian, Southeast Asian, White, Black, Hispanic, Middle Eastern, Indian), creating attribute-specific descriptions. Specifically, we add \textit{East Asian} for the East Asian class, \textit{Southeast Asian} for the Southeast Asian class, \textit{White} or \textit{Caucasian} for the White class, \textit{Black} or \textit{African} for the Black class, \textit{Latino}, \textit{Latina}, or \textit{Hispanic} for the Hispanic class, \textit{Middle Eastern} or \textit{Arab} for the Middle Eastern class, \textit{Indian} for the Indian class before the person-specific terms (\eg, \textit{person} $\rightarrow$ \{\textit{East Asian person}, \textit{Southeast Asian person}, \textit{White person}, \textit{Black person}, \textit{Latino person}, \textit{Middle Eastern person}, \textit{Indian person}\}).

\section{List of concepts and occupations}
\label{sec:concepts}
We show the concepts used to compute $\text{MaxSkew@$k$}$ (in Tab. 1 and 2 in the main paper). The list of the concepts for each category (\ie, adjectives, occupations, and activities) are as follows:

\vspace{3pt}
\noindent
\textbf{Adjectives:} \textit{kind}, \textit{successful}, \textit{noncriminal}, \textit{nonviolent}, \textit{moral}, \textit{rich}, \textit{honest}, \textit{smart}, \textit{evil}, \textit{unkind}, \textit{unsuccessful}, \textit{criminal}, \textit{violent}, \textit{immoral}, \textit{poor}, \textit{dishonest}, \textit{dumb}, \textit{clever}, \textit{stupid}, \textit{hardworking}, \textit{lazy}, \textit{nasty}, \textit{trustworthy}, \textit{untrustworthy}, \textit{caring}, \textit{dangerous}, \textit{heroic}, \textit{villainous}, \textit{good}. We refer these concepts from \citep{berg2022prompt}. Regardig the templates used to create prompts, we use ``This is the face of a \{\} person'', ``This is a photo of a \{\} person'', ``This person is \{\}'', ``A \{\} person'', ``A \{\} individual'', ``This individual is \{\}'', ``Photo of a \{\}'', ``\{\}'', ``This is a \{\}'', ``This is the photo of a \{\}'', ``A \{\}'', following \citep{berg2022prompt}. The curly brackets are replaced with each concept.

\vspace{3pt}
\noindent
\textbf{Occupations:} \textit{doctor}, \textit{aerospace engineer}, \textit{computer programmer}, \textit{electrical engineer}, \textit{scientist}, \textit{artist}, \textit{author}, \textit{designer}, \textit{musician}, \textit{painter}, \textit{photographer}, \textit{singer}, \textit{writer}, \textit{architect}, \textit{civil engineer}, \textit{engineer}, \textit{programmer}, \textit{software developer}, \textit{childcare worker}, \textit{coach}, \textit{dental assistant}, \textit{dental hygienist}, \textit{dentist}, \textit{clerk}, \textit{housekeeper}, \textit{maid}, \textit{massage therapist}, \textit{nurse}, \textit{therapist},\textit{psychologist}, \textit{social assistant}, \textit{social worker}, \textit{teacher}, \textit{professor}, \textit{CEO}, \textit{skateboarder}, \textit{surfer}, \textit{baseball player}, \textit{football player}, \textit{soccer player}, \textit{tennis player}. For the occupation list, we refer to \citep{friedrich2023fair} with some modifications, such as additional occupations for a more inclusive list. As for the prompt templates, we select ones that are suitable to occupations (\eg, ``Photo of a \{\}''). We use this list for the text-to-image experiments with some modifications. Specifically, we remove the similar occupations (\eg, remove \textit{civil engineer} as there is \textit{engineer}). Additionally, we remove occupations where the text-to-image model \citep{rombach2022high} can not generate images with humans or it tends to generate multiple individuals.

\vspace{3pt}
\noindent
\textbf{Activities:} arranging flowers, playing tennis, playing skateboarding, playing baseball, playing soccer, playing football, playing snowboarding, playing skiing, cleaning, dressmaking, tying tie, smiling, crying, laughing, cooking, making pizza, dancing, drinking beer, drinking wine, eating hotdog, eating cake, using computer, playing game, gardening, singing, petting dog, petting cat, makeup, shopping, playing piano, playing guitar, carrying baby. For activities, we use a subset of the Kinetics dataset \citep{kay2017kinetics} with some additional activities. For the prompt templates, we use ``This is the face of a person who likes \{\}'', `` This is a photo of a person who likes \{\}'', ``This person likes \{\}'', ``A person who likes \{\}'', ``Photo of a person who likes \{\}'', ``This is a person who likes \{\}''.


\section{Additional visual examples for image generation experiments}
\label{sec:examples}
\begin{figure}[ht]
  \centering
  \includegraphics[clip, width=0.99\textwidth]{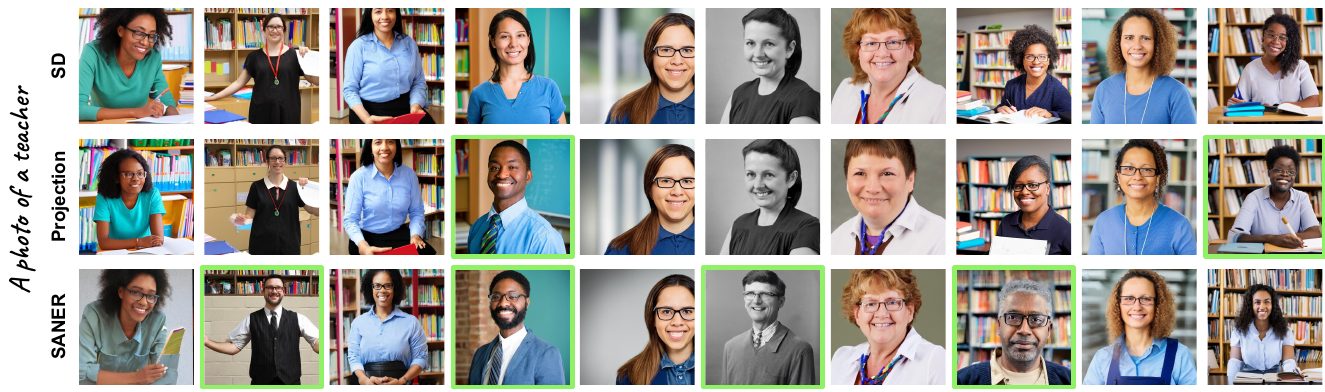}
  \vspace{-3pt}
  \caption{Generated images for the prompt, ``A photo of a teacher,'' by the original Stable Diffusion (SD), projection-based debiased CLIP (Projection), and our debiased CLIP (SANER). We randomly sample $10$ images from generated images. Images framed in \textcolor{colbest}{green} denote those of the minority gender in the generated images (\ie, male).}
  \label{fig:app-sd-bias}
  \vspace{-13pt}
\end{figure}

\begin{figure}[ht]
  \centering
  \includegraphics[clip, width=0.99\textwidth]{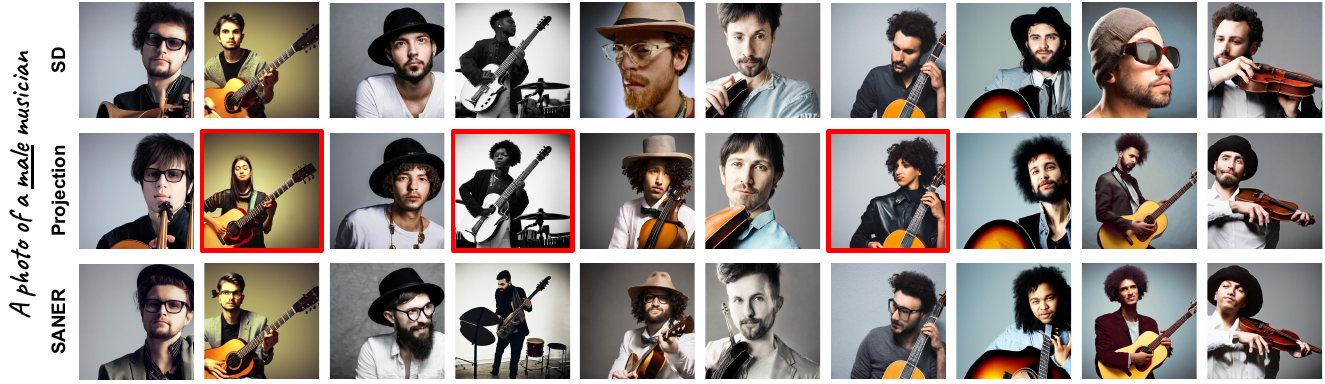}
  \vspace{-3pt}
  \caption{Generated images for the prompt, ``A photo of a male musician,'' by the original Stable Diffusion (SD), projection-based debiased CLIP (Projection), and our debiased CLIP (SANER). We randomly sample $10$ images from generated images. \textcolor{colworst}{Red} frame indicates images with incorrect gender (\ie, female).}
  \label{fig:app-sd-att}
  \vspace{-10pt}
\end{figure}

We show additional visual examples for Figures 3 and 4 in the main paper in Figures \ref{fig:app-sd-bias} and \ref{fig:app-sd-att}, respectively.

\section{Potential Extension}
\label{sec:extension}
\noindent \textbf{Additional Attributes. }
While we have aimed to make the attribute list as comprehensive as possible based on prior work \citep{berg2022prompt,seth2023dear,chuang2023debiasing}, we acknowledge that the choice of attribute groups may be subjective and that certain attributes might be missed. However, in contrast to previous approaches that require extensive attribute labels for images in the dataset, our method is designed to be flexible and extensible, allowing the attribute list to be expanded or adjusted based on specific needs or ethical considerations of the application domain.

\noindent
\textbf{Non-binary gender. }
While our experiments followed prior work \citep{berg2022prompt,seth2023dear,chuang2023debiasing} in focusing on binary gender, SANER naturally extends to non-binary gender by defining additional attribute-specific terms and their neutral forms (\eg, ``non-binary person'' $\rightarrow$ ``person''). The debiasing loss (Eq. \ref{eq:deb_loss}) can handle any number of attribute groups, making it straightforward to include non-binary gender categories.

\textbf{Automated text neutralization. }
While our current word-level approach is effective for well-defined social biases, it could be extended to handle more complex cases. As a potential future direction, embedding-based neutralization or context-aware language models could be explored to automate this process. These methods would enable bias identification and neutralization at a semantic level, reducing reliance on predefined attribute word lists and making the debiasing process more robust and adaptable to diverse scenarios.

\textbf{Debiasing LLaVA-like models. }
SANER can be extended to mitigate societal biases in large vision-language models, such as LLaVA \citep{liu2024visual}, by incorporating a debiasing mechanism for the image encoder. Specifically, this could involve training a debiasing layer to remove attribute information from the visual features, particularly for images without human subjects. We leave this extension as an avenue for future work.

\section{Potential Negative Impact}
\label{sec:impact}

Applying SANER to debias CLIP may lead to a potential negative impact where users might overlook remaining biases, assuming the process to be fully effective. While SANER performs better in gender and age bias mitigation, as evidenced by the MaxSkew and statistical parity metrics, this does not ensure that SANER mitigates all possible societal biases, and there could be dimensions of bias not adequately measured by these metrics. It is crucial to acknowledge that SANER, though impactful, is not an all-encompassing solution for removing societal bias. We notice that researchers must exercise due diligence in evaluating the application of SANER to avoid inadvertently introducing unanticipated biases.

\end{document}